\documentclass{article} %
\usepackage{paper,times}

\usepackage{amsmath,amsfonts,bm}

\def\eqref#1{equation~\ref{#1}}

\def\1{\bm{1}}

\DeclareMathAlphabet{\mathsfit}{\encodingdefault}{\sfdefault}{m}{sl}
\SetMathAlphabet{\mathsfit}{bold}{\encodingdefault}{\sfdefault}{bx}{n}

\usepackage{hyperref}
\usepackage{url}
\usepackage{graphicx}
\usepackage{flafter}
\usepackage{placeins}
\usepackage{wrapfig}
\usepackage{needspace}
\usepackage{booktabs}
\usepackage{xcolor}
\usepackage{array}
\usepackage{tabularx}
\usepackage{adjustbox}
\usepackage{cleveref}
\usepackage{dblfloatfix}
\usepackage{caption}
\usepackage{subcaption}
\usepackage{enumitem}
\title{VLRLA: Deployment-Time Adaptation of Vision-Language-Action Models through In-Context Reinforcement Learning}

\author{Diyuan Shi$^{1,2,\dagger}$, Yiqi Tang$^{2,\dagger}$, Hongyin Zhang$^{2,\dagger}$, You Wu$^{2}$, Donglin Wang$^{2,*}$ \\
$^{1}$Zhejiang University \quad $^{2}$Westlake University \quad $^{*}$Corresponding author \\
$^{\dagger}$Equal contribution
}

\iclrfinalcopy %
\begin{document}

\maketitle
\fancyhead{}

\begin{abstract}
    Vision-language-action (VLA) models connect visual perception and language understanding with robot control, providing a foundation for instruction-conditioned manipulation. However, conventional pretrained VLA policies lack a mechanism for gradual adaptation through accumulated interaction experience during deployment. We therefore propose VLRLA, a general framework that equips existing VLA models with \textit{in-context reinforcement learning} capabilities for deployment-time adaptation. Our approach augments a VLA model with a reinforcement learning module that learns from historical observations, actions, rewards, and termination signals. During deployment, the policy gradually refines its action generation and progressively improves task success while keeping model parameters fixed. Our approach is also designed for general applicability and compatibility with diverse learning history construction strategies and model training pipelines. Finally, through extensive experiments, we systematically investigate the learning properties of VLRLA in simulation and demonstrate its ability to adapt to unseen scenarios during real-world deployment.
\end{abstract}

\section{Introduction}
Vision-language-action (VLA) models connect visual observations and natural-language instructions to robot actions, enabling robots to interpret their surroundings and carry out user requests. Their potential applications span household assistance, warehouse logistics, and industrial manipulation \citep{collaborationOpenXEmbodimentRobotic2023,intelligence$p_05$VisionLanguageActionModel2025,luoPreciseDexterousRobotic2025,aiP07SteerableGeneralist,xuRLTokenBootstrapping2026}. %

Conventional VLA methods learn from collected demonstrations and could generalize to new situations. However, demonstration-based training alone does not provide a mechanism for gradual adaptation after deployment. A robot may encounter an unfamiliar camera viewpoint, lighting condition, or instruction phrasing, each of which hinders task performance. The central limitation is therefore the inability to turn accumulated deployment experience into adaptive behavior. %

Online Deep Reinforcement Learning (DRL) or interactive Imitation Learning (IIL) provide routes to deployment-time adaptation through parameter updates, using reward feedback or corrective actions, respectively~\citep{wagenmaker2025SteeringYourDiffusionPolicyLatentSpaceReinforcement,xuRLTokenBootstrapping2026,luoPreciseDexterousRobotic2025,intelligence$p^_06$VLAThat2025,panSOPScalableOnline2026}. However, these approaches have several drawbacks: 1) Collecting the samples needed for policy updates takes tens of minutes to hours and involves substantial human effort; 2) They typically require task-specific hyperparameter tuning; and 3) They update policy in each deployment environment, making it difficult to scale to multi-task settings. These motivate us to   explore an adaptation mechanism that uses deployment feedback without parameter updates.

In this work, we propose VLRLA, a framework for equipping VLA models with in-context Reinforcement Learning (RL) capabilities. The framework augments a VLA model with a RL module that learns to use past interactions for adaptation. During deployment, action generation is conditioned on a memory of previous interactions and reward feedback, while model parameters remain fixed. As experience accumulates, this memory informs refinements to subsequent behavior, helping policy adapt to the deployment environment. The intended result is gradual improvement in task success through continued interaction during deployment and no parameter updates.

Existing VLA works vary across training datasets, policy architectures, and deployment tasks, we aim to develop a \textit{general} VLRLA framework that accommodates various learning history construction strategies, policy instantiations, and integration strategies. %
Our contributions are as follows:
\begin{itemize}[leftmargin=1em,topsep=0pt]
    \item We formulate deployment-time adaptation in VLA models as an in-context reinforcement learning problem, using accumulated interactions and reward feedback to inform subsequent decisions while keeping model parameters fixed.
    \item We propose a \textit{general} framework that unifies learning history construction and adaptation-module integration to equip pretrained VLA models with in-context reinforcement learning capabilities across diverse data construction settings, model architectures, and deployment tasks.
    \item We conduct extensive experiments in simulation and on real-world robots to demonstrate the effectiveness of VLRLA. Our results show adaptation across multiple pretrained backbones and improved task success with accumulated experience, including on challenging unseen tasks.
\end{itemize}

\section{Related Work}
\textbf{Vision-Language-Action Models.} VLA models integrate visual and language representations with action prediction to support generalist robotic manipulation. Progress in this direction has centered on expanding the data and representations learning for training robot policies. The RT series~\citep{RT1RoboticsTransformer2022,brohanRT2VisionLanguageActionModels2023,collaborationOpenXEmbodimentRobotic2023} advances generalist robot control through multi-task learning, web-scale vision-language knowledge transfer, and training across robot embodiments. OpenVLA~\citep{kim2024OpenVLAOpenSourceVisionLanguageActionModela} makes VLA modeling accessible through an open-source model trained on diverse robot demonstrations and designed to support efficient fine-tuning. $\pi_{0.5}$~\citep{intelligence$p_05$VisionLanguageActionModel2025} pursues open-world generalization by co-training on heterogeneous robot and web data, combining semantic subtask prediction with robot control. LingBot-VLA~\citep{wuPragmaticVLAFoundation2026} further explores large-scale pretraining across multiple dual-arm robot configurations, releasing a model, training infrastructure, and evaluation data to support adaptation across tasks and platforms.  In this work, we study how to equip existing VLA models with in-context Reinforcement Learning capabilities for deployment-time adaptation. %

\textbf{Deep Reinforcement Learning.} DRL enables policies to improve through trial-and-error. Proximal Policy Optimization (PPO)~\citep{schulmanProximalPolicyOptimization2017} uses a policy-gradient surrogate objective to learn from on-policy rollouts. Soft Actor-Critic (SAC)~\citep{haarnojaSoftActorCriticAlgorithms2019} combines off-policy actor--critic learning with an entropy maximization framework for continuous control. Recent studies apply these policy optimization principles to finetune pretrained VLA models. Examining the generalization benefits of RL fine-tuning, \citet{liuWhatCanRL2026} report that PPO-based OpenVLA fine-tuning improves semantic and execution generalization over supervised fine-tuning.  SimpleVLA-RL~\citep{liSimpleVLARLScalingVLA2025} also combines parallel trajectory collection with VLA-specific training optimizations to scale reward-based RL for OpenVLA-OFT. Other methods focus on online DRL fine-tuning directly in real-world tasks: RLToken~\citep{xuRLTokenBootstrapping2026} uses a compact representation from a pretrained VLA for sample-efficient online DRL fine-tuning. Similarly, DSRL~\citep{wagenmaker2025SteeringYourDiffusionPolicyLatentSpaceReinforcement} learns an RL policy over the latent-noise input of a frozen diffusion or flow policy. These methods require parameter updates during RL fine-tuning, whereas our work studies in-context adaptation with all parameters fixed. %

\textbf{In-Context Learning.} Recent robotics research explores how demonstrations, prompts, and interaction histories can guide action generation at inference time for VLA models. For adaptation from demonstrations, In-Context Robot Transformer (ICRT)~\citep{fuInContextImitationLearning2024} predicts actions from sensorimotor trajectories, allowing new tasks to be specified through demonstration prompts without parameter fine-tuning. $\pi_{0.7}$~\citep{aiP07SteerableGeneralist} combines language, subgoal images, and episode metadata to steer task execution, highlighting the role of multimodal context in generalization. In concurrent work, RoboTTT~\citep{jiangRoboTTTContextScaling2026} compresses visuomotor context into recurrent fast weights through test-time training, focusing on one-shot imitation from human videos and online corrections learned through DAgger~\citep{rossReductionImitationLearning2011} distillation. Specifically, reward-driven adaptation is addressed by Algorithm Distillation (AD)~\citep{laskinIncontextReinforcementLearning2022} which performs in-context RL in discrete simulation environments. \citet{huang2026incontextlearningrobotsmethods} provides a survey about in-context learning on robots. In contrast, our work focuses on designing a general framework to equip existing VLA models with in-context RL capabilities across diverse learning history construction strategies and model settings with validated effectiveness in both simulation and real-world continuous robot control tasks.

\section{Preliminaries}
\label{sec:preliminaries}

\textbf{Vision-Language-Action Policy.} At timestep $t$, a VLA model $\pi$ receives a language instruction $\ell$ and an observation $o_t=(I_t,q_t)$ where  $I_t$ is camera image and  $q_t$ is robot proprioceptive state, then generates an action chunk
\begin{equation}
 \pi(o_t, \ell) \rightarrow a_{t:t+h} \in\mathbb{R}^{h \times d_a}
\end{equation}
where $h$ is the action-chunk length, $d_a$ is the dimension of each action, and $a_{t:t+h}$ contains the $h$ actions $a_t,\ldots,a_{t+h-1}$, excluding the endpoint $t+h$. The robot controller then executes the action chunk before replanning. 

\textbf{In-Context Reinforcement Learning.} A \textit{trial} is a sequence of interactions that may span \textit{multiple episodes} of the same task.  We define the available history as $\mathcal{C}_t=\{(o_i,a_{i:i+h},r_i,d_i)\}_{i=0}^{t-1}$, with $\mathcal{C}_0=\emptyset$, where $r_i$ is the observed reward and $d_i$ marks an episode boundary. This history contains only completed interactions before step $t$. An in-context RL policy $a_{t:t+h} \sim \pi_\psi( \cdot \mid o_t,\ell,\mathcal{C}_t)$ performs action generation  while keeping $\psi$ fixed. For tasks $\mathcal{M}\sim p(\mathcal{M})$ and a discount factor $\gamma \in [0,1)$, the learning objective is to optimize the task return:
\begin{equation}
    J(\psi)=\mathbb{E}_{\mathcal{M}\sim p(\mathcal{M}),\,\pi_\psi}
    \biggl[\sum_{t\ge0}\gamma^t r_t\biggr].
    \label{eq:contextual-return}
\end{equation}

\textbf{Algorithm Distillation (AD)}. Algorithm Distillation (AD), introduced by \citet{laskinIncontextReinforcementLearning2022}, trains a causal sequence model on ordered learning histories $\mathcal{C}\sim\mathcal{D}_{\mathrm{hist}}$ collected from a source RL algorithm:
\begin{equation}
    \mathcal{L}_{\mathrm{AD}}(\psi)
    =-\mathbb{E}_{\mathcal{C}\sim\mathcal{D}_{\mathrm{hist}}}
    \left[\sum_{t=0}
    \log\pi_\psi(a_t\mid o_t,\ell,\mathcal{C}_t)\right].
    \label{eq:ad-objective}
\end{equation}
Preserving the order of learning across episodes helps the model learn how action generation changes with experience. At deployment, the model could perform in-context RL and gradually improve the task return by conditioning on its own interaction history in \textit{unseen} tasks, without parameter updates. %

\begin{figure}[t]
    \centering
    \includegraphics[width=\textwidth]{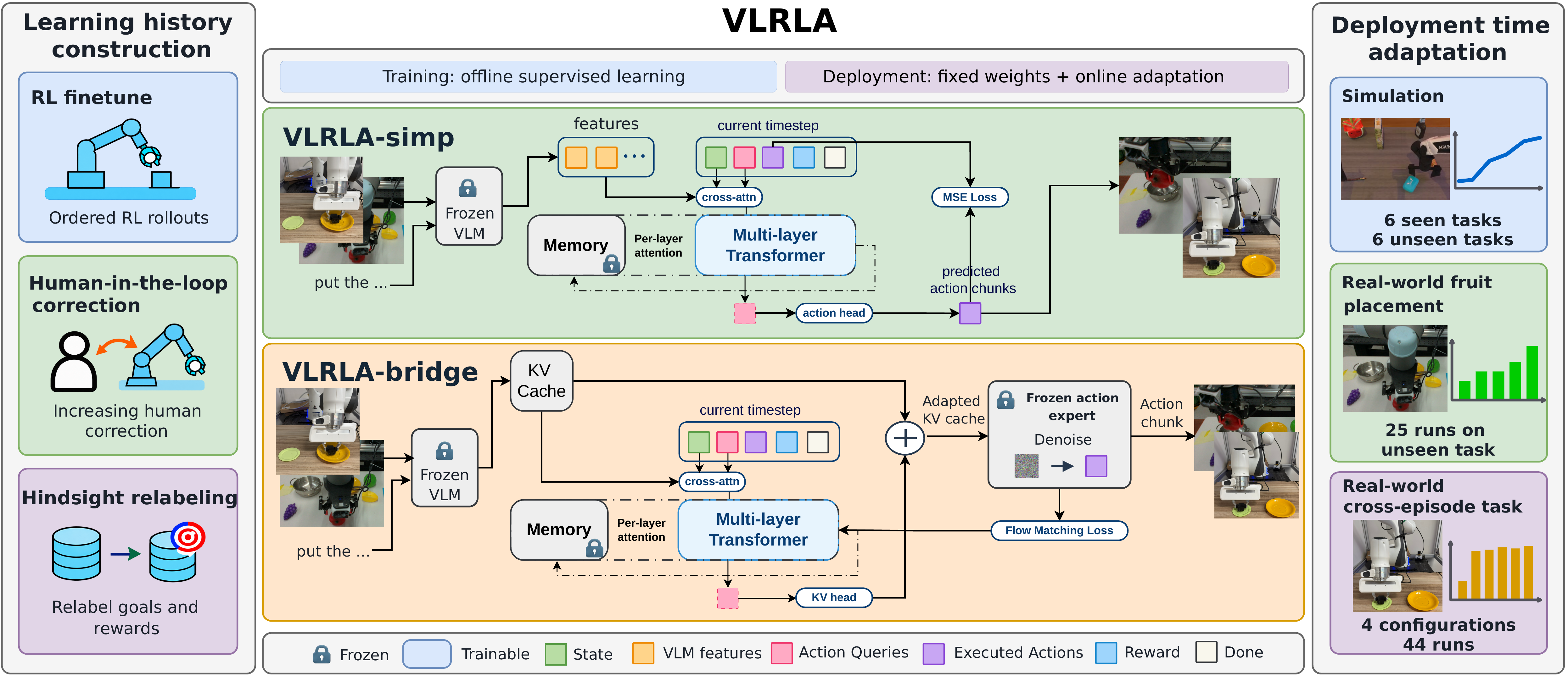}
    \caption{Framework overview. VLRLA is designed as a general in-context RL framework supporting diverse learning history construction strategies and model integration strategies.}
    \label{fig:framework}
\end{figure}

\section{Training Adaptive Vision-Language-Action Models}
\label{sec:method}
Algorithm Distillation (AD) provides a principled approach to distilling RL algorithms into causal sequence models, but its integration with existing pretrained VLA models remains unexplored. VLA methods differ in their data collection procedures, training pipelines, and deployment settings. We therefore design VLRLA as a general framework and demonstrate its effectiveness across multiple settings. This section presents the VLRLA framework, illustrated in \Cref{fig:framework}.  %

\subsection{Model Design}
Deployment-time adaptation requires a robotic policy to integrate visual and language information with interaction feedback, retain experience over long horizons, and generate continuous actions. Our design addresses these requirements through recurrent memory, pretrained visual-language representations, and two policy instantiations that connect these representations to action generation.

\textbf{Recurrent Memory.} In robotic tasks, a trial can span multiple episodes and thousands of tokens. To reduce the substantial memory and computation costs of dot-product causal attention~\citep{vaswaniAttentionAllYou2017}, we use recurrent memory similar to Transformer-XL~\citep{daiTransformerXLAttentiveLanguage2019}. This mechanism combines causal attention within a local window of timesteps with attention to cached representations from earlier windows. The policy forward pass is expressed as
\begin{equation}
    (M_t, \hat{a}_{t:t+h}) = \pi_\psi(o_t, \ell, a_{t-1:t-1+h}, r_{t-1}, d_{t-1} \mid M_{t-1}),
    \label{eq:recurrent-memory}
\end{equation}
where $\pi_\psi$ denotes the full policy with parameters $\psi$, $M_{t-1}$ denotes memory from the preceding step, and the outputs are the updated memory $M_t$ and generated action chunk $\hat{a}_{t:t+h}$. Memory retains information across episodes within a trial, and model parameters remain fixed during deployment.

\textbf{Visual and Language Preprocessing.}
\label{sec:method-architecture}
We reuse representations from a pretrained VLA model to avoid training a visual-language encoder from scratch. The framework accommodates two interfaces: the encoder's prefix embeddings and its key--value (KV) cache. Both provide information about the current observation and instruction for action generation, while recurrent memory supplies context from previous interactions.
For the prefix-embedding interface, we extract visual-language embeddings and pass them to the recurrent policy:
\begin{equation}
    \begin{aligned}
        z_t^{\mathrm{emb}} &= f_{\bar{\theta}_{\mathrm{VLM}}}^{\mathrm{emb}}(o_t,\ell),\\
        (M_t,\hat{a}_{t:t+h}) &= \pi_{\phi}^{\mathrm{emb}}(q_t,z_t^{\mathrm{emb}},a_{t-1:t-1+h},r_{t-1},d_{t-1}\mid M_{t-1}),
    \end{aligned}
    \label{eq:prefix-policy}
\end{equation}
where $f_{\bar{\theta}_{\mathrm{VLM}}}^{\mathrm{emb}}$ is the pretrained encoder that extracts prefix embeddings $z_t^{\mathrm{emb}}$ from the current observation $o_t$ and instruction $\ell$, and $\pi_{\phi}^{\mathrm{emb}}$ is the downstream recurrent policy. The bar over $\theta_{\mathrm{VLM}}$ indicates that the encoder is frozen during training, while $\phi$ denotes the policy parameters.

For the KV-cache interface, we condition the recurrent policy on the attention keys and values extracted from the VLM:
\begin{equation}
    \begin{aligned}
        z_t^{\mathrm{KV}} &= f_{\bar{\theta}_{\mathrm{VLM}}}^{\mathrm{KV}}(o_t,\ell),\\
        (M_t,\hat{a}_{t:t+h}) &= \pi_{\phi}^{\mathrm{KV}}(q_t,z_t^{\mathrm{KV}},a_{t-1:t-1+h},r_{t-1},d_{t-1}\mid M_{t-1}),
    \end{aligned}
    \label{eq:kv-policy}
\end{equation}
where $f_{\bar{\theta}_{\mathrm{VLM}}}^{\mathrm{KV}}$ extracts the VLM's KV cache $z_t^{\mathrm{KV}}$ and $\pi_{\phi}^{\mathrm{KV}}$ denotes the recurrent policy. The KV cache exposes layer-specific attention keys and values across multiple VLM layers, providing access to intermediate representations for action generation.

\textbf{Policy Instantiations.} Building on these memory and representation interfaces, we introduce two VLRLA variants for different task settings. \texttt{VLRLA-simp} uses a deterministic policy intended for simple tasks or simulation environments. \texttt{VLRLA-bridge} retains the pretrained visual-language encoder and stochastic action expert and is intended for complex tasks or real-world deployment.

\texttt{VLRLA-simp}: We pair a frozen visual-language encoder with a trainable recurrent action module (RLA). The RLA consists of a Transformer with recurrent memory and a multilayer perceptron (MLP) that produces deterministic actions, as illustrated in the upper part of \Cref{fig:framework}:
\begin{equation}
    \begin{aligned}
        z_t &= f_{\bar{\theta}_{\mathrm{VLM}}}(o_t, \ell),\\
        (M_t, \hat{a}_{t:t+h}) &= \pi_{\theta_{\mathrm{RLA}}}(q_t, z_t, a_{t-1:t-1+h}, r_{t-1}, d_{t-1} \mid M_{t-1}),
    \end{aligned}
    \label{eq:vla-approach1}
\end{equation}
where $z_t$ denotes the prefix embeddings $z_t^{\mathrm{emb}}$ introduced above, and $q_t$ is the proprioceptive state. The RLA parameters $\theta_{\mathrm{RLA}}$ are trainable, while the encoder parameters $\bar{\theta}_{\mathrm{VLM}}$ remain frozen. %
We train this deterministic policy by minimizing a reconstruction loss:
\begin{equation}
    \mathcal{L}(\theta_{\mathrm{RLA}})
    =\mathbb{E}_{\mathcal{C}\sim\mathcal{D}}
    \Bigl[\sum_{t=0}
    \mathrm{MSE}\bigl(\hat{a}_{t:t+h}, a_{t:t+h}\bigr)\Bigr].
\end{equation}
where $\mathrm{MSE}$ denotes mean squared error and $a_{t:t+h}$ is the target action chunk in the dataset.  %

\texttt{VLRLA-bridge}: Our second variant reuses both the pretrained visual-language encoder and the action expert. A trainable adaptation module connects these frozen components by modifying the encoder's KV cache, as illustrated in the lower part of \Cref{fig:framework}. 
Using the $\pi_{0.5}$ architecture as an example, the policy takes the form
\begin{equation}
    \begin{aligned}
        z_t &= f_{\bar{\theta}_{\mathrm{VLM}}}(o_t, \ell),\\
        (M_t, \Delta z_t) &= f_{\theta_{\mathrm{RL}}}(q_t, z_t, a_{t-1:t-1+h}, r_{t-1}, d_{t-1} \mid M_{t-1}),\\
        z_t' &= z_t + \Delta z_t,
        \qquad a_{t:t+h} \sim \pi_{\bar{\theta}_{\mathrm{A}}}(\cdot\mid q_t,z_t'),
    \end{aligned}
    \label{eq:vla-approach2}
\end{equation}
where $z_t$ denotes the KV cache $z_t^{\mathrm{KV}}$ introduced above. The bridge, with trainable parameters $\theta_{\mathrm{RL}}$, produces a residual $\Delta z_t$ with the same structure as $z_t$, yielding the adapted cache $z_t'$. The frozen action expert, parameterized by $\bar{\theta}_{\mathrm{A}}$, generates actions conditioned on the adapted cache and the proprioceptive state. Through this residual modification, interaction history influences the expert's action generation. Further implementation details are provided in \Cref{app:architecture}.

The pretrained $\pi_{0.5}$ action expert generates continuous action chunks through iterative flow-based sampling. We therefore train \texttt{VLRLA-bridge} with the corresponding flow-matching objective, propagating gradients through the frozen expert to the bridge:
\begin{equation}
    \mathcal{L}(\theta_{\mathrm{RL}})
    =\mathbb{E}_{\mathcal{C}\sim\mathcal{D},\,\epsilon,\,\tau}
    \Bigl[\bigl\|v_{\bar{\theta}_{\mathrm{A}}}(a_{t:t+h}^\tau,\tau,q_t,z_t') - (\epsilon-a_{t:t+h})\bigr\|_2^2\Bigr].
\end{equation}
where $a_{t:t+h}^\tau=(1-\tau)a_{t:t+h}+\tau\epsilon$ is the interpolated noisy action chunk, $\epsilon\sim\mathcal{N}(0,I)$ is Gaussian noise, and $\tau\in[0,1]$ is the interpolation time.  Although the expert is frozen, gradients backpropagate to $\theta_{\mathrm{RL}}$ through $z_t'$. %

\subsection{Learning History Construction}
\label{sec:method-data}
Training VLRLA requires learning histories that capture how behavior improves with accumulated interaction experience. To accommodate different data sources and collection settings, our framework supports three strategies for constructing the training dataset: simulation RL, curriculum-based human-in-the-loop collection, and hindsight-based assembly. Details are in \Cref{app:dataset}

\textbf{Simulation RL.} Following AD~\citep{laskinIncontextReinforcementLearning2022a}, we collect trajectories throughout an RL training process in simulation. This approach can use algorithms such as PPO and GRPO~\citep{shaoDeepSeekMathPushingLimits2024} to fine-tune base VLA models such as Pi05, LingbotVLA, and FastWAM~\cite{yuanFastWAMWorldAction2026}. Unlike AD, we find that resetting the environment to the same initial state after every episode termination or truncation is unnecessary and departs from standard practice in robotic tasks that use domain randomization~\citep{daiRoboMMEBenchmarkingUnderstanding2026,chenRoboTwin20Scalable2025,chenRoboDojoUnifiedSimandReal2026}. Instead, in each training round, we sample a batch of environment instances through domain randomization within the same task, such as \textit{``picking up diverse bottles''}. We then collect on-policy trajectories with the current VLA model, store them as part of the training dataset, and perform an online RL update to obtain the next policy checkpoint. We repeat this process for several rounds. The resulting dataset captures successive rounds of interaction as task performance improves and is used to train VLRLA to transfer the source RL algorithm's adaptation behavior to unseen tasks.
\label{method:sim_rl}

\textbf{Human-in-the-Loop Curriculum.} Human-in-the-loop (HITL) correction is widely used in robot learning to improve robot behavior. Starting from an existing VLA policy, we construct a training dataset by progressively increasing the extent of human intervention. We first collect trajectories generated entirely by the base VLA policy, without human correction. In the next stage, a human operator uses teleoperation to correct the first $1/3$ of each episode, after which the VLA policy completes the remainder. We then increase the fraction of each episode under human intervention from $1/3$ to $2/3$ and finally to $3/3$, with the final stage consisting almost entirely of human demonstrations. This curriculum produces a dataset of progressively improved behavior, ordered from unassisted policy rollouts through partial corrections to human expert trajectories. These sequences can be used to train VLRLA to adapt to unseen environments and gradually improve task performance.

\textbf{Hindsight Relabeling.} Hindsight experience replay (HER)~\citep{andrychowiczHindsightExperienceReplay2017} reuses trajectories by relabeling them with alternative goals. For a language-conditioned task, let $\ell_1$ and $\ell_2$ denote two language instructions, and let $R_\ell$ denote the instruction-conditioned reward function. A transition can be relabeled as follows:
\begin{equation}
    (o_t,a_t,R_{\ell_1}(o_t,a_t,o_{t+1}),o_{t+1},\ell_1)
    \longmapsto
    (o_t,a_t,R_{\ell_2}(o_t,a_t,o_{t+1}),o_{t+1},\ell_2).
    \label{eq:her-relabel}
\end{equation}
For example, a trajectory that places a power adapter in a green plate receives a reward of 1 under the instruction \textit{``put the power adapter into the green plate''} but a reward of 0 under \textit{``put the power adapter into the yellow plate''}. This relabeling allows us to assemble training histories without additional data collection: for the task \textit{``put the power adapter into one of the plates''} (more in \Cref{sect:rw_2}), we can collect successful demonstrations of placement in either the green or yellow plate and use HER relabeling to assemble histories of progressively adaptive behavior.

\section{Simulation Experiments}
\label{sec:simulation}
In this part, we conduct simulation experiments to study the following research questions: 1) How is \texttt{VLRLA} compared to other algorithms? 2) Can VLRLA learn to adapt using different VLM backbones?  3) How does context length affect \texttt{VLRLA}'s ability to adapt?  4) Could the history of one rollout be transferred? We use \texttt{VLRLA-simp} in 2, 3 and 4 for its simplicity and better reproducibility and use \texttt{VLRLA-bridge} in 1 as its pretrained modules could benefit the multi-task learning most.

We use the RoboTwin benchmark~\citep{chenRoboTwin20Scalable2025} as our testbed for its diverse and challenging tasks, and RLinf~\citep{yuRLinfFlexibleEfficient2025} to collect simulation data through online RL (\textit{Simulation RL} in \Cref{sec:method-data}). We compute success rates by evaluating policies across 128 randomized environment instances and averaging the outcomes.

\begin{table}[h!]
    \centering
    \small
    \setlength{\tabcolsep}{3pt}
    \renewcommand{\arraystretch}{1.08}
    \caption{Algorithm characteristics and success rates (\%) on seen and unseen RoboTwin tasks. $\Delta$ denotes the improvement of VLRLA-bridge over $\pi_{0.5}$ in percentage points. ``/'' indicates \textit{not applicable}. The highest success rate in each row is shown in bold, including all ties.}
    \label{tab:sim-task-summary}
    \begin{tabularx}{\textwidth}{@{}>{\raggedright\arraybackslash}X|>{\centering\arraybackslash}p{0.07\textwidth}|>{\centering\arraybackslash}p{0.13\textwidth}|>{\centering\arraybackslash}p{0.11\textwidth}>{\centering\arraybackslash}p{0.23\textwidth}|>{\raggedleft\arraybackslash}p{0.09\textwidth}@{}}
        \toprule
         & $\pi_{0.5}$ & GRPO & RoboTTT & VLRLA-bridge & $\Delta$ \\
        \midrule
        \textbf{In-context adaptation} & No & No & Yes & Yes & / \\
        \textbf{Sample efficiency} & / & Low & Moderate & Moderate & / \\
        \midrule
        \multicolumn{6}{@{}l}{\textbf{Seen tasks}} \\
        \texttt{handover\_mic} & 38.3\% & 32.8\% & 52.3\% & \textbf{78.9\%} & 40.6\% \\
        \texttt{pick\_diverse\_bottles} & 58.6\% & \textbf{77.3\%} & 71.1\% & 75.8\% & 17.2\% \\
        \texttt{place\_a2b\_left} & 67.2\% & \textbf{81.2\%} & 70.3\% & 71.9\% & 4.7\% \\
        \texttt{place\_object\_stand} & 71.1\% & 87.5\% & 76.6\% & \textbf{90.6\%} & 19.5\% \\
        \texttt{place\_phone\_stand} & 36.7\% & \textbf{50.8\%} & 47.7\% & \textbf{50.8\%} & 14.1\% \\
        \texttt{scan\_object} & 19.5\% & 25.8\% & \textbf{28.9\%} & \textbf{28.9\%} & 9.4\% \\
        \midrule
        \textit{Seen average} & 48.6\% & 59.2\% & 57.8\% & \textbf{66.2\%} & 17.6\% \\
        \midrule
        \multicolumn{6}{@{}l}{\textbf{Unseen tasks}} \\
        \texttt{adjust\_bottle} & 95.3\% & \textbf{100.0\%} & \textbf{100.0\%} & \textbf{100.0\%} & 4.7\% \\
        \texttt{move\_stapler\_pad} & 24.2\% & \textbf{38.3\%} & 31.2\% & 36.7\% & 12.5\% \\
        \texttt{place\_a2b\_right} & 61.7\% & \textbf{78.1\%} & 69.5\% & 69.5\% & 7.8\% \\
        \texttt{place\_dual\_shoes} & 17.2\% & \textbf{21.1\%} & 19.5\% & \textbf{21.1\%} & 3.9\% \\
        \texttt{place\_empty\_cup} & 93.0\% & \textbf{99.2\%} & 97.7\% & 97.7\% & 4.7\% \\
        \texttt{stamp\_seal} & 28.1\% & 28.1\% & 33.6\% & \textbf{38.3\%} & 10.2\% \\
        \midrule
        \textit{Unseen average} & 53.3\% & \textbf{60.8\%} & 58.6\% & 60.6\% & 7.3\% \\
        \midrule
        \textbf{Overall average} & 50.9\% & 60.0\% & 58.2\% & \textbf{63.4\%} & 12.4\% \\
        \bottomrule
\end{tabularx}
\end{table}

\subsection{How does \texttt{VLRLA} compare with other algorithms?}
\label{sec:sim-task-summary}
We first compare VLRLA with baseline methods on seen and unseen RoboTwin tasks. We evaluate three categories of algorithms:
\begin{itemize}[leftmargin=1em,topsep=0pt]
    \item Supervised fine-tuning: $\pi_{0.5}$ fine-tuned with supervised learning serves as the baseline. This model also serves as the base model for below methods.
    \item RL fine-tuning: GRPO~\citep{shaoDeepSeekMathPushingLimits2024}. We use the official implementation provided by RLinf. To support on-policy data collection in parallel environments, our training setup requires 8 NVIDIA A100-80G GPUs.
    \item In-context adaptation: our \texttt{VLRLA-bridge} and RoboTTT~\citep{jiangRoboTTTContextScaling2026}. We train both methods on the same multi-task dataset collected by fine-tuning $\pi_{0.5}$ on six seen tasks, but omit reward and termination information  to RoboTTT.  During deployment, both methods adapt with accumulated interaction histories while reward and termination are omitted again for RoboTTT. We use our own implementation of RoboTTT, as its official code has not been released.
\end{itemize}

As shown in \Cref{tab:sim-task-summary}, \texttt{VLRLA-bridge} improves over the initial $\pi_{0.5}$ policy on all 12 tasks, with an average gain of 12.4 percentage points. Its overall average success rate exceeds that of GRPO, an online RL method that explicitly updates policy parameters through stochastic gradient descent. Compared with RoboTTT, \texttt{VLRLA-bridge} also achieves a higher overall average success rate. This gap reflects the absence of reward signals in RoboTTT hinders the model's adaptation capability. On the six \textit{seen tasks}, it improves over $\pi_{0.5}$ by an average of 17.6 percentage points and achieves a higher average success rate than GRPO.

On the six \textit{unseen tasks}, \texttt{VLRLA-bridge} improves over $\pi_{0.5}$ on every task, with an average gain of 7.3 percentage points. Its average success rate reaches $60.6\%$, approaching that of GRPO. It also achieves a higher average success rate than RoboTTT on unseen tasks, outperforming it on three tasks and matching its success rate on the remaining three.  These results suggest that supervised pretraining on learning histories from seen tasks enables the model to learn an adaptation capability that generalizes to unseen tasks.

\begin{wrapfigure}[11]{r}{0.50\textwidth}
    \centering
    \vspace{-10pt}
    \includegraphics[width=\linewidth]{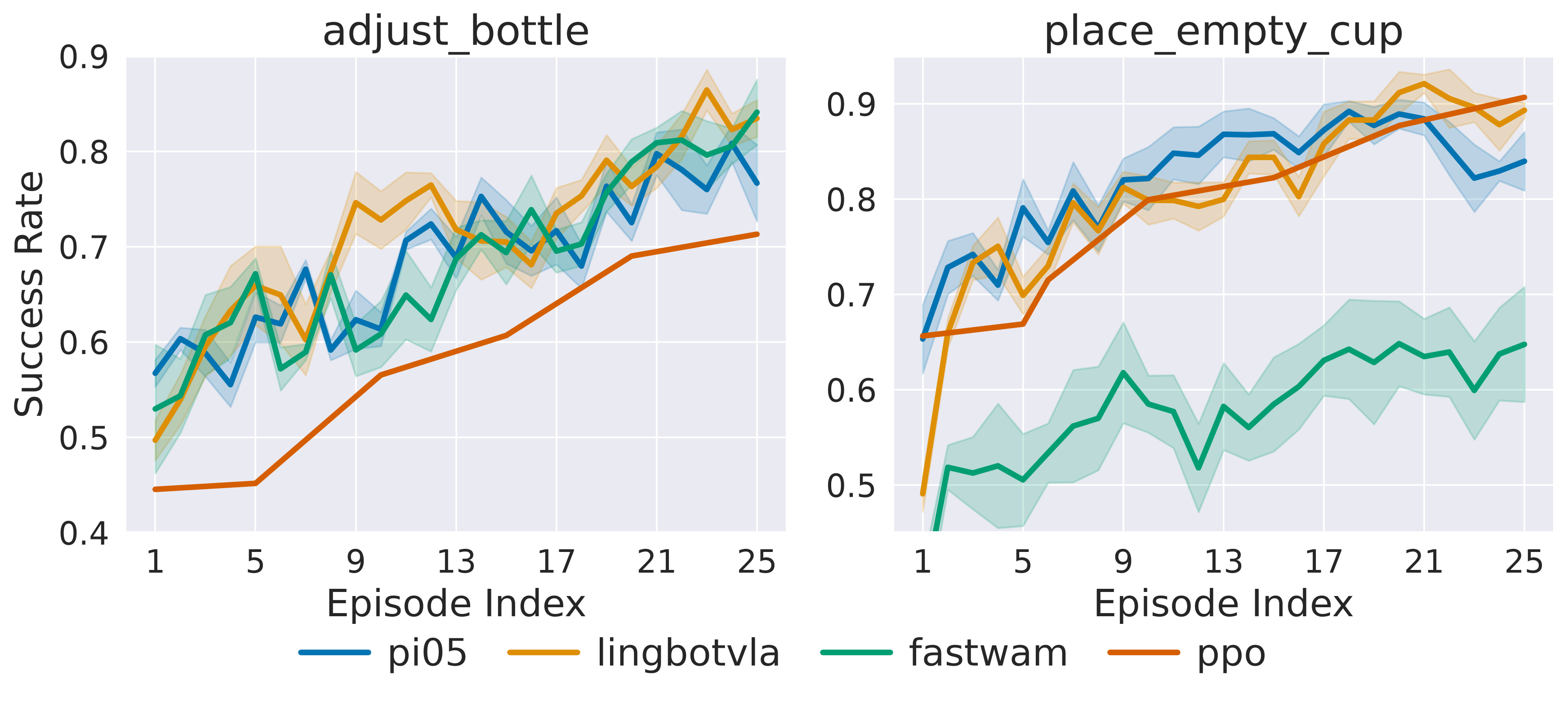}
    \caption{Adaptation across VLM modules. Shaded area is  1 standard deviation across 5 seeds.}
    \label{fig:sim-base-models}
\end{wrapfigure}

\subsection{Can VLRLA learn to adapt using different VLM backbones?}
\label{sec:sim-base-models}
\Cref{fig:sim-base-models} evaluates whether \texttt{VLRLA-simp} can learn to adapt using representations from different pretrained backbones: $\pi_{0.5}$, \texttt{lingbotvla}, and the world action model \texttt{fastwam}. All three configurations improve as interaction experience accumulates, supporting the framework's applicability across different representation backbones. For both \texttt{lingbotvla} and $\pi_{0.5}$, success rates generally increase with accumulated interaction history on both tasks. Although \texttt{fastwam} also learns to adapt, its success rate on \texttt{place\_empty\_cup} remains lower than those of the other two backbones. We attribute this performance gap to the constrained prefix output used in our FastWAM integration, which limits the visual and language information available to the adaptation and policy module (see the \Cref{app:fastwam} for details). Together, these results suggest that \texttt{VLRLA-simp} can learn to adapt across different backbones and highlight the importance of strong visual-language representations. %

\begin{wrapfigure}[11]{r}{0.50\textwidth}
    \centering
    \vspace{-16pt}
    \includegraphics[width=\linewidth]{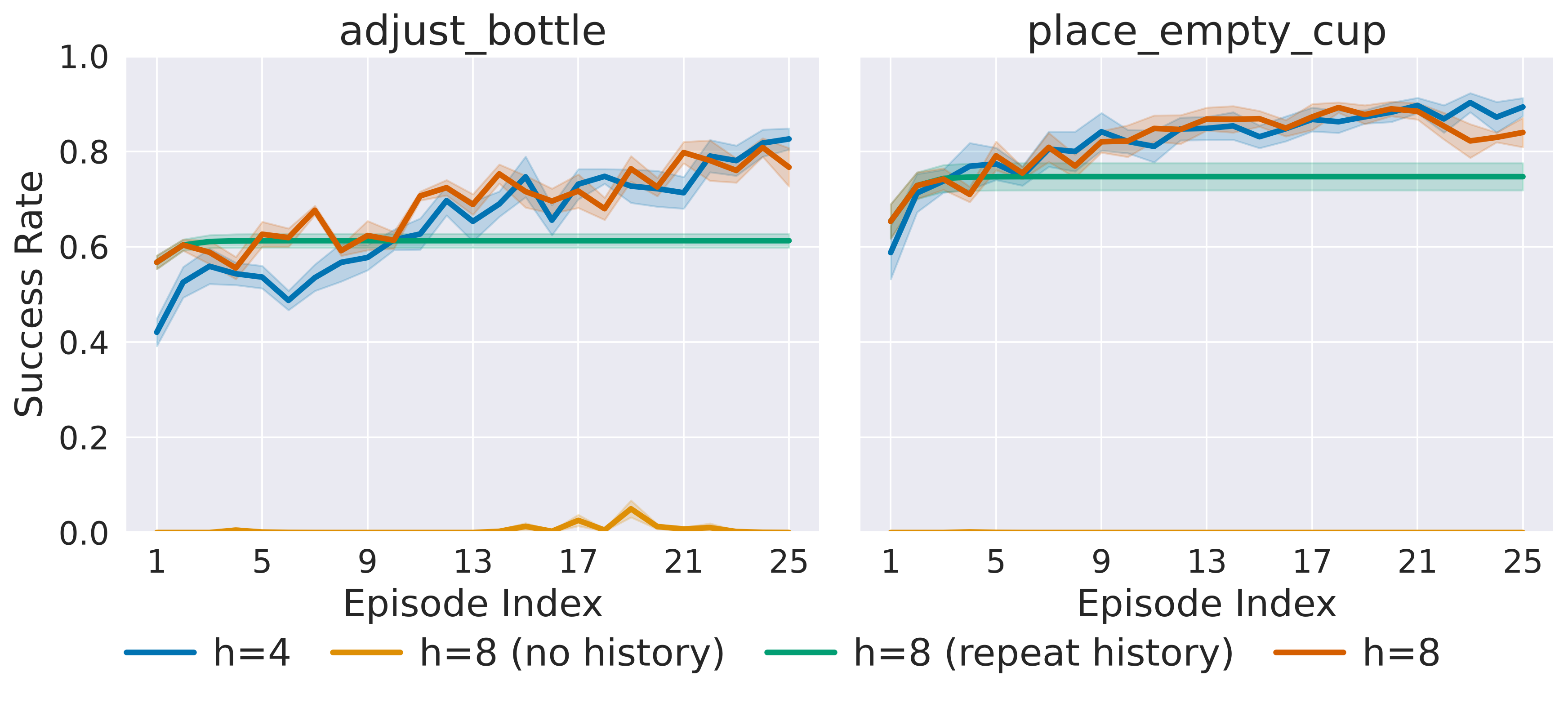}
    \caption{Effect of context and memory.  Shaded area is  1 standard deviation across 5 seeds.}
    \label{fig:sim-h}
\end{wrapfigure}
\subsection{How does context length affect \texttt{VLRLA-simp}'s ability to adapt? }
\label{sec:sim-h}
\Cref{fig:sim-h} examines how context length and interaction history affect adaptation. With context length $h=4$ and $h=8$, \texttt{VLRLA-simp} achieves broadly similar performance on both \texttt{adjust\_bottle} and \texttt{place\_empty\_cup}, with success rates gradually improving as interaction experience accumulates. In the $h=8$ (no history) setting, we explicitly zero out the history memory when generating action chunks for the current episode and observe a substantial drop in success rate: performance remains near zero, with no sustained improvement across episodes. We also evaluate an $h=8$ (repeat history) variant that reuses the first episode's history at every subsequent step. The success rate improves at the first step after this history becomes available, then remains flat because no new interaction information is incorporated. Together, these experiments highlight the importance of both access to interaction history and its continued accumulation for adaptation. %

\begin{wrapfigure}[12]{r}{0.50\textwidth}
    \centering
    \vspace{-14pt}
    \includegraphics[width=\linewidth]{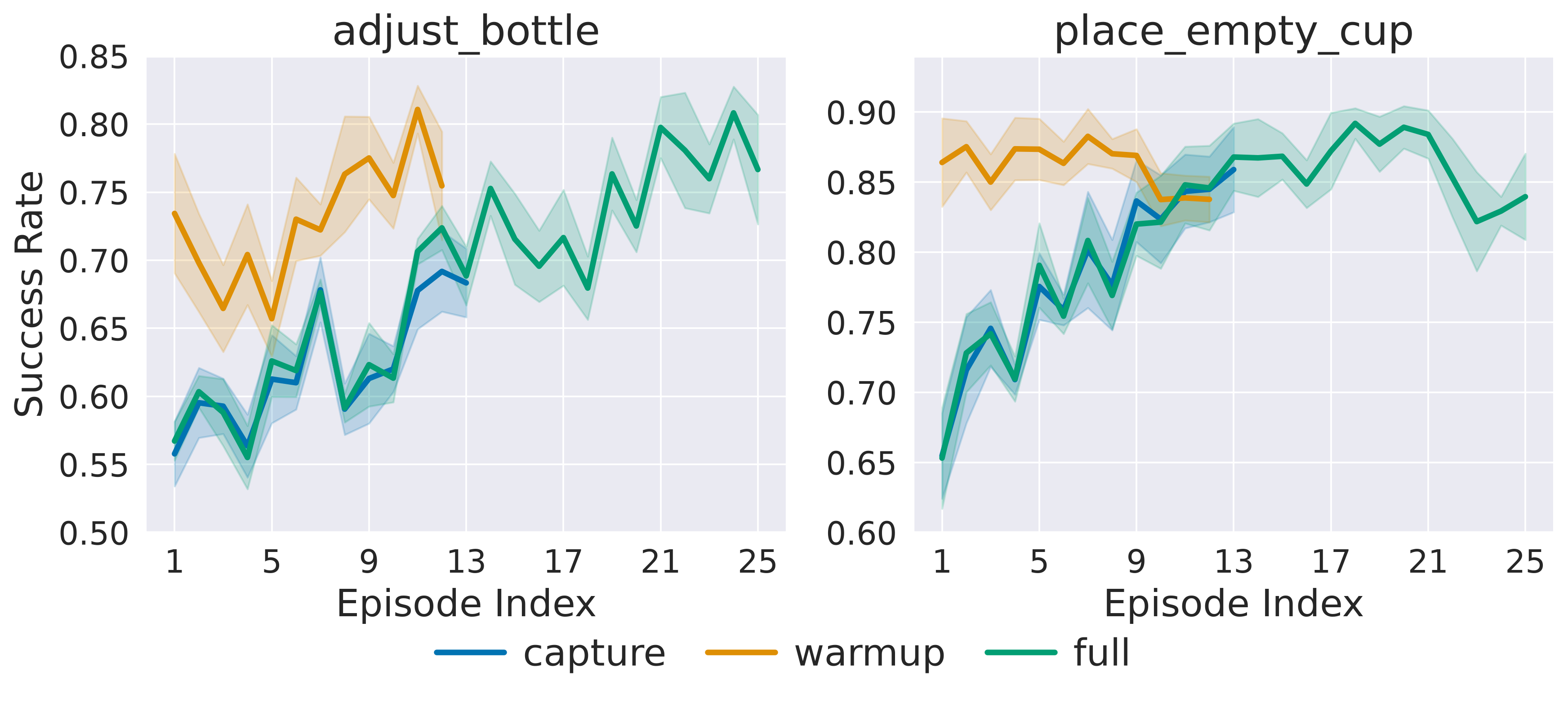}
    \caption{History transfer to improve initial task performance. Shaded regions indicate one standard deviation across five seeds.}
    \label{fig:sim-warmup}
\end{wrapfigure}
\subsection{Could the history of one rollout be transferred?}
\label{sec:sim-warmup}
\Cref{fig:sim-warmup} examines whether interaction history collected in one rollout can improve initial performance in another rollout of the same task with different initial states or configurations. With this history loaded, the \texttt{warmup} rollout begins with higher success rates on both tasks: approximately $0.7$ versus $0.55$ for the \texttt{capture} and \texttt{full} rollouts on \texttt{adjust\_bottle}, and $0.85$ versus $0.65$ on \texttt{place\_empty\_cup}. The \texttt{capture} rollout supplies the interaction history used by \texttt{warmup}, while \texttt{full} denotes the original rollout without transferred history. These results show that interaction history can be transferred between rollouts of the same task to improve performance in a new configuration.

\section{Real World Experiments}
\label{sec:real-world}
\begin{figure}[htbp]
    \centering
    \includegraphics[width=\textwidth]{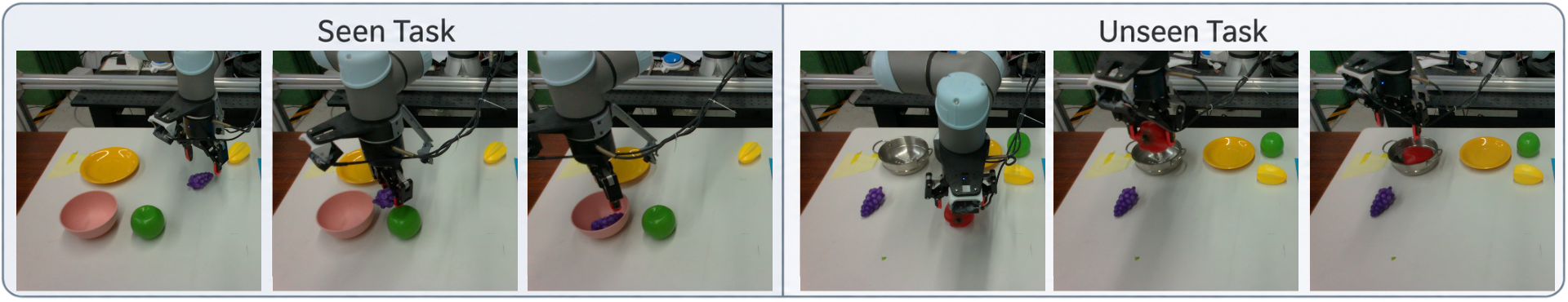}
    \caption{Left 3 figures depict the fruits placement. Right 3 figures depict the pepper placement.}
    \label{fig:rw-fruit-demo}
    \vspace{-12pt}
\end{figure}
\subsection{Fruit-Placement}
\label{sec:rw-aggregate}
\begin{wrapfigure}[23]{r}{0.36\textwidth}
    \centering
    \vspace{-34pt}
    \begin{subfigure}{\linewidth}
        \centering
        \includegraphics[width=\linewidth]{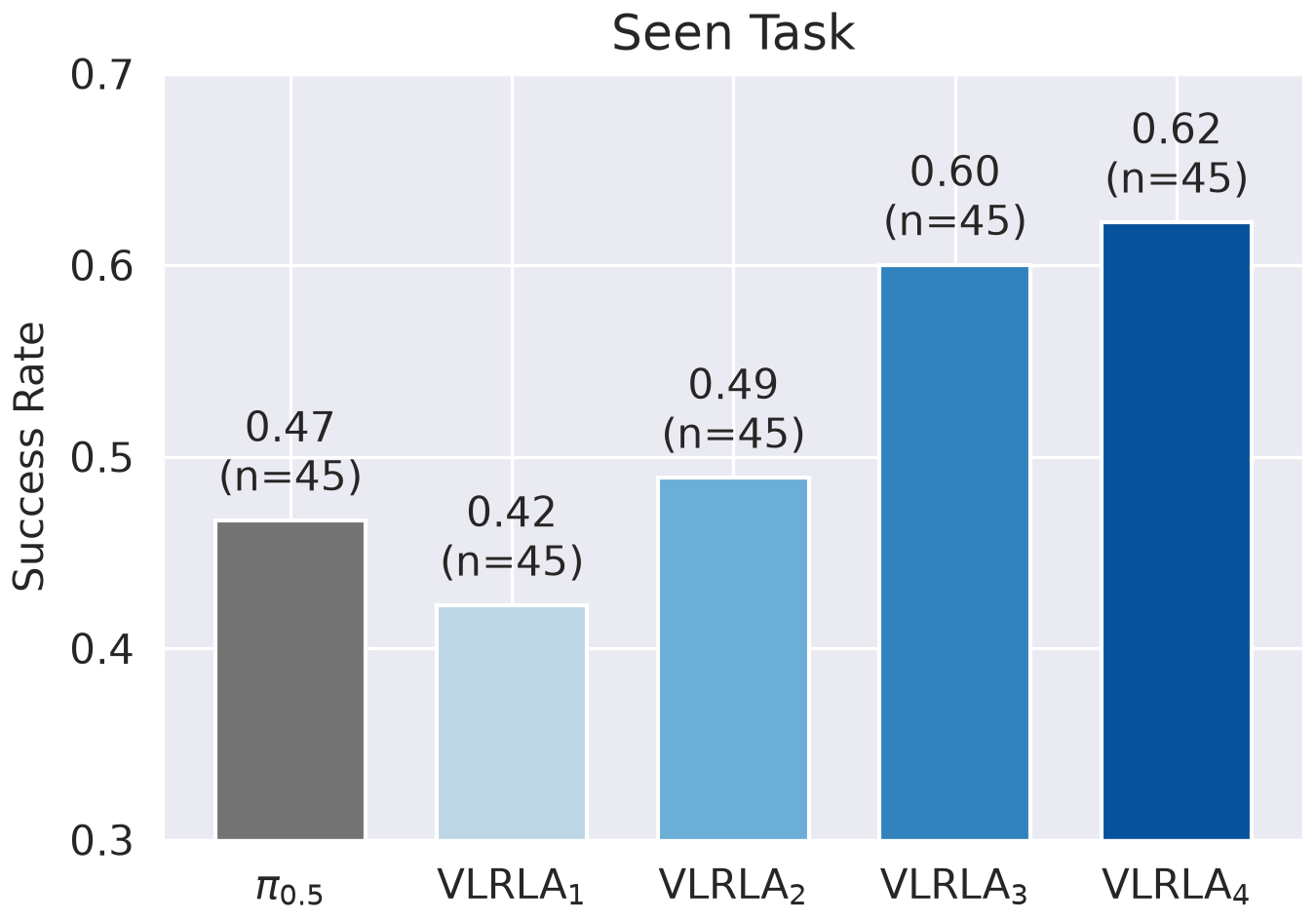}
        \caption{Results for seen fruit task averaged in 45 runs.}
        \label{fig:rw-aggregate}
    \end{subfigure}
    \par\medskip
    \begin{subfigure}{\linewidth}
        \centering
        \includegraphics[width=\linewidth]{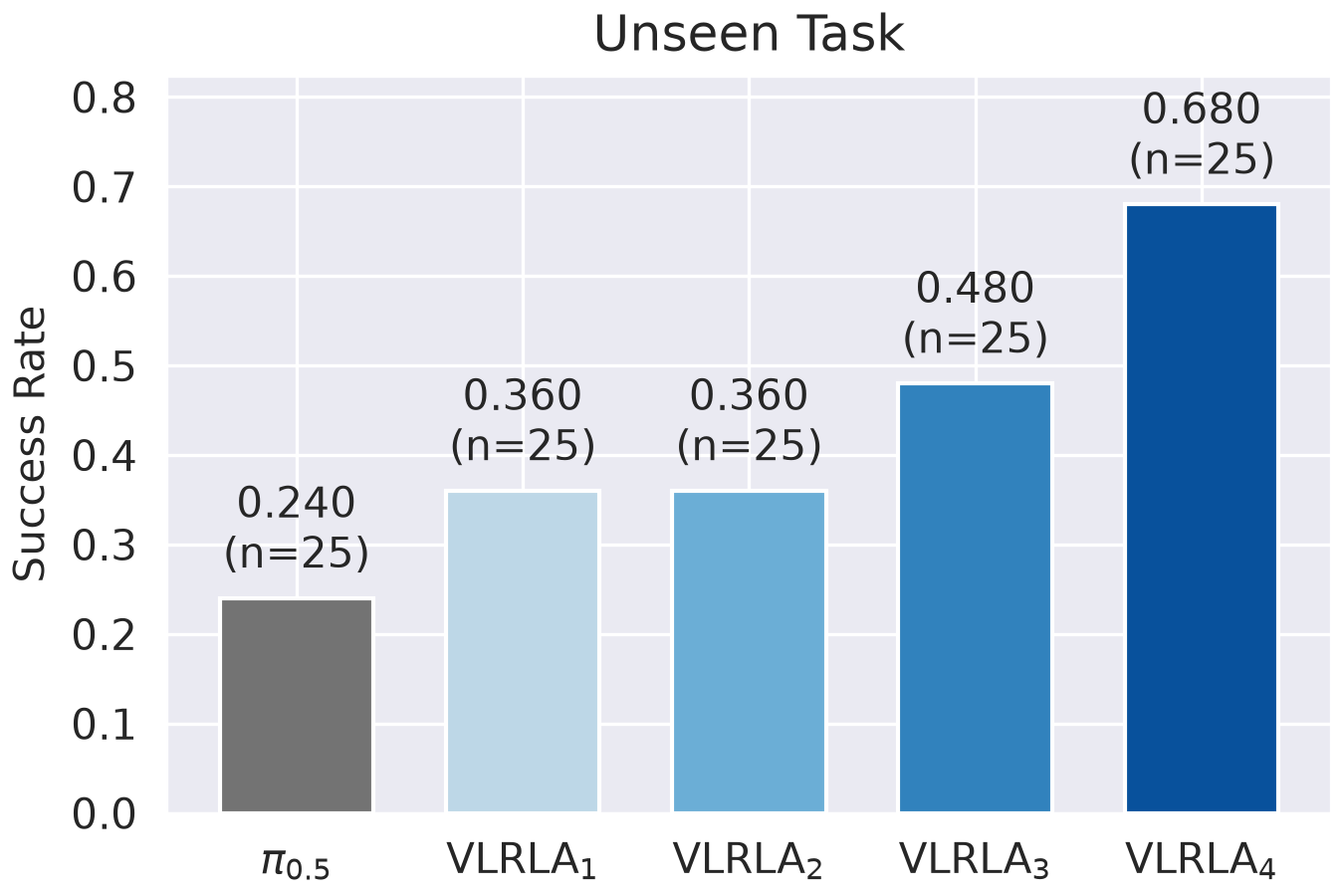}
        \caption{Result for unseen fruit task in 25 runs.}
        \label{fig:rw-pepper}
    \end{subfigure}
    \caption{Success rates on fruit-placement tasks.}
    \label{fig:rw-placement-results}
\end{wrapfigure}
This task requires the robot arm to pick up a fruit and place it in a vessel as in \Cref{fig:rw-fruit-demo}. The instruction is \textit{``put the \{fruit\} into \{vessel\}''}. The fruits used during training are \textit{apple}, \textit{starfruit}, and \textit{grape}; the vessels are \textit{bowl} and \textit{plate}. We evaluate whether the success rate of \texttt{VLRLA-bridge} improves as experience accumulates across these fruit--vessel combinations. We also evaluate transfer to an \textit{unseen setting} with a \textit{red pepper} and a \textit{metal pot}, neither of which appears in the training dataset. The dataset contains approximately 80 teleoperation episodes (used to fine-tune the $\pi_{0.5}$ base model), followed by $3\times20$ episodes with HITL corrections (described in \textit{Human-in-the-Loop Curriculum}, \Cref{sec:method-data}).  The videos for this task are in Supplemental Materials.

\Cref{fig:rw-aggregate} reports fruit-placement performance across 45 evaluation runs. Here, $\mathrm{VLRLA}_{1}$ -- $\mathrm{VLRLA}_{4}$ denote \texttt{VLRLA-bridge} evaluated over four successive episodes with accumulating interaction history, where $\mathrm{VLRLA}_{1}$ corresponds to the first episode with no prior interaction history. After a slight initial decline relative to $\pi_{0.5}$, VLRLA steadily improves with accumulated interactions and surpasses the baseline, demonstrating its ability to use accumulated experience to improve fruit-placement performance. 

For the unseen task \textit{``put the pepper into the pot,''} we use a medium-sized red pepper whose asymmetric shape makes it more challenging to grasp. \Cref{fig:rw-pepper} reports performance across 25 evaluation runs. VLRLA outperforms the $\pi_{0.5}$ baseline steadily. These results indicate that VLRLA can use accumulated experience to improve performance even when neither the object nor the vessel was encountered during training.

\Cref{fig:rw-fruit-breakdown} presents results by fruit type, comparing $\pi_{0.5}$ with $\mathrm{VLRLA}_{1}$--$\mathrm{VLRLA}_{4}$. For apple and starfruit, success rates increase steadily as interaction experience accumulates and exceed the $\pi_{0.5}$ baseline at $\mathrm{VLRLA}_{2}$. Performance on grape follows a less consistent trend, fluctuating across adaptation stages. Closer inspection suggests that this difference arises from the grape's small size and softness, combined with the relatively high force applied by the gripper. These properties make grape grasping less sensitive to the grasp location, limiting the benefit of adaptation. Apple and starfruit, however, require more precise grasp placement, leaving greater room for improvement.
\begin{table}[!htbp]
    \centering
    \small
    \caption{Fruit-placement success-rate breakdown by fruit type, with 15 runs per fruit. }
    \label{fig:rw-fruit-breakdown}
    \begin{tabular}{@{}l*{5}{c}>{\bfseries\boldmath}r@{}}
        \toprule
        & $\pi_{0.5}$ & $\mathrm{VLRLA}_{1}$ & $\mathrm{VLRLA}_{2}$ & $\mathrm{VLRLA}_{3}$ & $\mathrm{VLRLA}_{4}$ & $\mathrm{VLRLA}_{4}-\pi_{0.5}$ \\
        \midrule
        Apple & 46.7\% & 40.0\% & 53.3\% & 53.3\% & 60.0\% & +13.3\% \\
        Starfruit & 40.0\% & 26.7\% & 46.7\% & 60.0\% & 66.7\% & +26.7\% \\
        Grape & 53.3\% & 60.0\% & 46.7\% & 66.7\% & 60.0\% & +6.7\% \\
        \bottomrule
    \end{tabular}
\end{table}
\subsection{Cross-Episode Memory Task}
\label{sect:rw_2}
\begin{figure}[!htbp]
    \centering
    \includegraphics[width=0.93\textwidth]{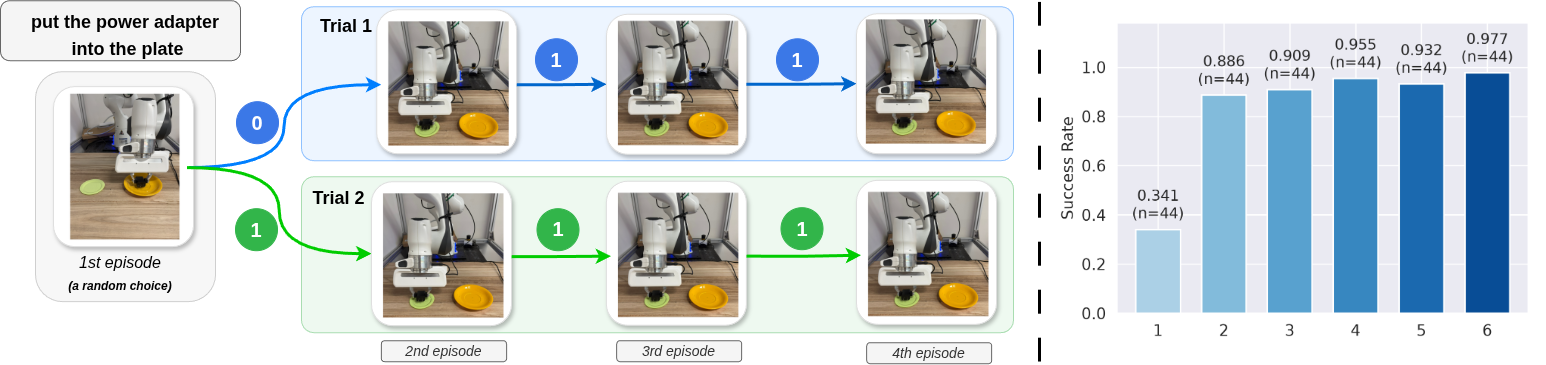}
    \caption{Left part: illustrative diagram for cross-episodic-memory task. The digit in circle denotes the reward label passed to policy after the episode. Right part: success rate averaged in 44 runs.}
    \label{fig:rw-cross-episodic}
\end{figure}

In this task, we examine whether \texttt{VLRLA-bridge} can adapt its behavior using reward feedback from previous episodes. A green plate, a yellow plate, and a power adapter are placed on a table, and the policy receives the underspecified instruction \textit{``put the power adapter into the plate''}. The target plate's color determines the reward but is deliberately withheld from the policy. Placing the adapter in the target plate yields a reward of~1, while any other outcome yields a reward of~0. The left  of \Cref{fig:rw-cross-episodic} illustrates the task setup. This task tests whether the policy can employ reward feedback across episode boundaries and use it to guide subsequent plate choices. We use \textit{hindsight relabeling} to dynamically assemble the dataset using only expert demonstrations for \textit{``put the power adapter into the green plate''} and \textit{``put the power adapter into the yellow plate''} (details in \Cref{app:mem_data}).

The right panel of \Cref{fig:rw-cross-episodic} shows a steady increase in success rate after the first episode, when reward feedback becomes available. Performance quickly rises in subsequent episodes, suggesting that the policy can use reward feedback from earlier episodes to guide later choices. Videos of this task are available in the supplementary material.

\section{Conclusion}
VLRLA enables deployment-time adaptation through in-context reinforcement learning, conditioning action generation on accumulated interactions, rewards, and termination signals while keeping model parameters fixed. Our general framework accommodates multiple strategies for constructing learning histories and policy instantiations.

Through extensive experiments in simulation and real-world settings, we demonstrate that VLRLA can improve task success using accumulated interaction experience, including challenging unseen tasks. Simulation studies show adaptation with different pretrained backbones and highlight the importance of continually updating interaction history. Real-world experiments further show improved manipulation performance and suggest that reward feedback retained across episodes can inform subsequent action choices.

\bibliography{iclr2027_conference}

\begin{thebibliography}{30}
\providecommand{\natexlab}[1]{#1}
\providecommand{\url}[1]{\texttt{#1}}
\expandafter\ifx\csname urlstyle\endcsname\relax
  \providecommand{\doi}[1]{doi: #1}\else
  \providecommand{\doi}{doi: \begingroup \urlstyle{rm}\Url}\fi

\bibitem[RT1(2022)]{RT1RoboticsTransformer2022}
{{RT-1}}: {{Robotics Transformer}} for real-world control at scale.
\newblock https://ai.googleblog.com/2022/12/rt-1-robotics-transformer-for-real.html, December 2022.

\bibitem[Ai et~al.()Ai, Amin, Aniceto, Balakrishna, Balke, Black, Bokinsky, Cao, Charbonnier, Choudhary, Collins, Conley, Connors, Darpinian, Dhabalia, Dhaka, DiCarlo, Driess, Equi, Esmail, Fang, Finn, Glossop, Godden, Goryachev, Groom, Habeeb, Hancock, Hausman, Hussein, Hwang, Ichter, Jacobsen, Jakubczak, Jen, Jones, Kammerer, Katz, Ke, Khadikov, Kuchi, Lamb, LeBlanc, LeCount, Levine, Li, {Li-Bell}, Lialin, Liang, Lim, Lu, Luo, Mano, Marwaha, Mongush, Murphy, Nair, Patterson, Pertsch, Ren, Schelske, Sharma, Shi, Shi, Smith, Springenberg, Stachowicz, Stoeckle, Tang, Tanner, Tekeste, Torne, Vedder, Vuong, Walling, Wang, Wang, Wang, Whalen, Whitmore, Williams, Xu, Yoo, Yu, Zhang, Zhang, and Zhilinsky]{aiP07SteerableGeneralist}
Bo~Ai, Ali Amin, Raichelle Aniceto, Ashwin Balakrishna, Greg Balke, Kevin Black, George Bokinsky, Shihao Cao, Thomas Charbonnier, Vedant Choudhary, Foster Collins, Ken Conley, Grace Connors, James Darpinian, Karan Dhabalia, Maitrayee Dhaka, Jared DiCarlo, Danny Driess, Michael Equi, Adnan Esmail, Yunhao Fang, Chelsea Finn, Catherine Glossop, Thomas Godden, Ivan Goryachev, Lachlan Groom, Haroun Habeeb, Hunter Hancock, Karol Hausman, Gashon Hussein, Victor Hwang, Brian Ichter, Connor Jacobsen, Szymon Jakubczak, Rowan Jen, Tim Jones, Gregg Kammerer, Ben Katz, Liyiming Ke, Mairbek Khadikov, Chandra Kuchi, Marinda Lamb, Devin LeBlanc, Brendon LeCount, Sergey Levine, Xinyu Li, Adrian {Li-Bell}, Vladislav Lialin, Zhonglin Liang, Wallace Lim, Yao Lu, Enyu Luo, Vishnu Mano, Nandan Marwaha, Aikys Mongush, Liam Murphy, Suraj Nair, Tyler Patterson, Karl Pertsch, Allen~Z Ren, Gavin Schelske, Charvi Sharma, Baifeng Shi, Lucy~Xiaoyang Shi, Laura Smith, Jost~Tobias Springenberg, Kyle Stachowicz, Will Stoeckle, Jiaming Tang, Jimmy Tanner, Shalom Tekeste, Marcel Torne, Kyle Vedder, Quan Vuong, Anna Walling, Haohuan Wang, Jason Wang, XuDong Wang, Chris Whalen, Samuel Whitmore, Blake Williams, Charles Xu, Sukwon Yoo, Lili Yu, Wuming Zhang, Zhuoyang Zhang, and Ury Zhilinsky.
\newblock {$\pi$}0.7: A {{Steerable Generalist Robotic Foundation Model}} with {{Emergent Capabilities}}.

\bibitem[Andrychowicz et~al.(2017)Andrychowicz, Wolski, Ray, Schneider, Fong, Welinder, McGrew, Tobin, Abbeel, and Zaremba]{andrychowiczHindsightExperienceReplay2017}
Marcin Andrychowicz, Filip Wolski, Alex Ray, Jonas Schneider, Rachel Fong, Peter Welinder, Bob McGrew, Josh Tobin, Pieter Abbeel, and Wojciech Zaremba.
\newblock Hindsight experience replay.
\newblock In \emph{Proceedings of the 31st {{International Conference}} on {{Neural Information Processing Systems}}}, {{NIPS}}'17, pp.\  5055--5065, Red Hook, NY, USA, December 2017. Curran Associates Inc.
\newblock ISBN 978-1-5108-6096-4.

\bibitem[Brohan et~al.(2023)Brohan, Brown, Carbajal, Chebotar, Chen, Choromanski, Ding, Driess, Dubey, Finn, Florence, Fu, Arenas, Gopalakrishnan, Han, Hausman, Herzog, Hsu, Ichter, Irpan, Joshi, Julian, Kalashnikov, Kuang, Leal, Lee, Lee, Levine, Lu, Michalewski, Mordatch, Pertsch, Rao, Reymann, Ryoo, Salazar, Sanketi, Sermanet, Singh, Singh, Soricut, Tran, Vanhoucke, Vuong, Wahid, Welker, Wohlhart, Wu, Xia, Xiao, Xu, Xu, Yu, and Zitkovich]{brohanRT2VisionLanguageActionModels2023}
Anthony Brohan, Noah Brown, Justice Carbajal, Yevgen Chebotar, Xi~Chen, Krzysztof Choromanski, Tianli Ding, Danny Driess, Avinava Dubey, Chelsea Finn, Pete Florence, Chuyuan Fu, Montse~Gonzalez Arenas, Keerthana Gopalakrishnan, Kehang Han, Karol Hausman, Alexander Herzog, Jasmine Hsu, Brian Ichter, Alex Irpan, Nikhil Joshi, Ryan Julian, Dmitry Kalashnikov, Yuheng Kuang, Isabel Leal, Lisa Lee, Tsang-Wei~Edward Lee, Sergey Levine, Yao Lu, Henryk Michalewski, Igor Mordatch, Karl Pertsch, Kanishka Rao, Krista Reymann, Michael Ryoo, Grecia Salazar, Pannag Sanketi, Pierre Sermanet, Jaspiar Singh, Anikait Singh, Radu Soricut, Huong Tran, Vincent Vanhoucke, Quan Vuong, Ayzaan Wahid, Stefan Welker, Paul Wohlhart, Jialin Wu, Fei Xia, Ted Xiao, Peng Xu, Sichun Xu, Tianhe Yu, and Brianna Zitkovich.
\newblock {{RT-2}}: {{Vision-Language-Action Models Transfer Web Knowledge}} to {{Robotic Control}}, July 2023.

\bibitem[Chen et~al.(2025)Chen, Chen, Chen, Cai, Liu, Li, Liang, Lin, Ge, Gu, Deng, Guo, Nian, Xie, Chen, Su, Xu, Liu, Hu, Gao, Wang, Liang, Qin, Yang, Luo, and Mu]{chenRoboTwin20Scalable2025}
Tianxing Chen, Zanxin Chen, Baijun Chen, Zijian Cai, Yibin Liu, Zixuan Li, Qiwei Liang, Xianliang Lin, Yiheng Ge, Zhenyu Gu, Weiliang Deng, Yubin Guo, Tian Nian, Xuanbing Xie, Qiangyu Chen, Kailun Su, Tianling Xu, Guodong Liu, Mengkang Hu, Huan-ang Gao, Kaixuan Wang, Zhixuan Liang, Yusen Qin, Xiaokang Yang, Ping Luo, and Yao Mu.
\newblock {{RoboTwin}} 2.0: {{A Scalable Data Generator}} and {{Benchmark}} with {{Strong Domain Randomization}} for {{Robust Bimanual Robotic Manipulation}}, August 2025.

\bibitem[Chen et~al.(2026)Chen, Chen, Li, Tang, Su, Lu, Wan, Chen, Liu, Yan, Su, Dou, Wang, Zhang, Liu, Qin, Liang, Wu, Lin, Lin, Wang, He, Wu, Wu, Zhou, Lei, Yu, Ji, Jin, Lin, Li, Xiong, Xu, Li, Chai, Xie, Wang, Mu, Dong, Matusik, Ding, Ding, Luo, and Tomizuka]{chenRoboDojoUnifiedSimandReal2026}
Tianxing Chen, Yue Chen, Zixuan Li, Junyuan Tang, Kailun Su, Haoran Lu, Weijie Wan, Baijun Chen, Songling Liu, Haowen Yan, Honghao Su, Zhiyang Dou, Kaixuan Wang, Dandan Zhang, Yunze Liu, Yan Qin, Qiwei Liang, Qiwei Wu, Zijian Lin, Wenwei Lin, Yuran Wang, Minghua He, Tianshu Wu, Ruihai Wu, Jingquan Zhou, Kai-Chong Lei, Haibao Yu, Yuanfeng Ji, Weiyang Jin, Guanyu Lin, Xiaofan Li, Qi~Xiong, Renjing Xu, Zhongyu Li, Wenhao Chai, Enze Xie, Ziwei Wang, Yao Mu, Hao Dong, Wojciech Matusik, Mingyu Ding, Wenbo Ding, Ping Luo, and Masayoshi Tomizuka.
\newblock {{RoboDojo}}: {{A Unified Sim-and-Real Benchmark}} for {{Comprehensive Evaluation}} of {{Generalist Robot Manipulation Policies}}, July 2026.

\bibitem[Collaboration et~al.(2023)Collaboration, Padalkar, Pooley, Jain, Bewley, Herzog, Irpan, Khazatsky, Rai, Singh, Brohan, Raffin, Wahid, {Burgess-Limerick}, Kim, Sch{\"o}lkopf, Ichter, Lu, Xu, Finn, Xu, Chi, Huang, Chan, Pan, Fu, Devin, Driess, Pathak, Shah, B{\"u}chler, Kalashnikov, Sadigh, Johns, Ceola, Xia, Stulp, Zhou, Sukhatme, Salhotra, Yan, Schiavi, Kahn, Su, Fang, Shi, Amor, Christensen, Furuta, Walke, Fang, Mordatch, Radosavovic, Leal, Liang, {Abou-Chakra}, Kim, Peters, Schneider, Hsu, Bohg, Bingham, Wu, Wu, Luo, Gu, Tan, Oh, Malik, Tompson, Yang, Lim, Silv{\'e}rio, Han, Rao, Pertsch, Hausman, Go, Gopalakrishnan, Goldberg, Byrne, Oslund, Kawaharazuka, Zhang, Rana, Srinivasan, Chen, Pinto, Tan, Ott, Lee, Tomizuka, Du, Ahn, Zhang, Ding, Srirama, Sharma, Kim, Kanazawa, Hansen, Heess, Joshi, Suenderhauf, Di~Palo, Shafiullah, Mees, Kroemer, Sanketi, Wohlhart, Xu, Sermanet, Sundaresan, Vuong, Rafailov, Tian, Doshi, {Mart{\'i}n-Mart{\'i}n}, Mendonca, Shah, Hoque, Julian, Bustamante, Kirmani, Levine, Moore, Bahl, Dass, Sonawani, Song, Xu, Haldar, Adebola, Guist, Nasiriany, Schaal, Welker, Tian, Dasari, Belkhale, Osa, Harada, Matsushima, Xiao, Yu, Ding, Davchev, Zhao, Armstrong, Darrell, Jain, Vanhoucke, Zhan, Zhou, Burgard, Chen, Wang, Zhu, Li, Lu, Chebotar, Zhou, Zhu, Xu, Wang, Bisk, Cho, Lee, Cui, Wu, Tang, Zhu, Li, Iwasawa, Matsuo, Xu, and Cui]{collaborationOpenXEmbodimentRobotic2023}
Open X.-Embodiment Collaboration, Abhishek Padalkar, Acorn Pooley, Ajinkya Jain, Alex Bewley, Alex Herzog, Alex Irpan, Alexander Khazatsky, Anant Rai, Anikait Singh, Anthony Brohan, Antonin Raffin, Ayzaan Wahid, Ben {Burgess-Limerick}, Beomjoon Kim, Bernhard Sch{\"o}lkopf, Brian Ichter, Cewu Lu, Charles Xu, Chelsea Finn, Chenfeng Xu, Cheng Chi, Chenguang Huang, Christine Chan, Chuer Pan, Chuyuan Fu, Coline Devin, Danny Driess, Deepak Pathak, Dhruv Shah, Dieter B{\"u}chler, Dmitry Kalashnikov, Dorsa Sadigh, Edward Johns, Federico Ceola, Fei Xia, Freek Stulp, Gaoyue Zhou, Gaurav~S. Sukhatme, Gautam Salhotra, Ge~Yan, Giulio Schiavi, Gregory Kahn, Hao Su, Hao-Shu Fang, Haochen Shi, Heni~Ben Amor, Henrik~I. Christensen, Hiroki Furuta, Homer Walke, Hongjie Fang, Igor Mordatch, Ilija Radosavovic, Isabel Leal, Jacky Liang, Jad {Abou-Chakra}, Jaehyung Kim, Jan Peters, Jan Schneider, Jasmine Hsu, Jeannette Bohg, Jeffrey Bingham, Jiajun Wu, Jialin Wu, Jianlan Luo, Jiayuan Gu, Jie Tan, Jihoon Oh, Jitendra Malik, Jonathan Tompson, Jonathan Yang, Joseph~J. Lim, Jo{\~a}o Silv{\'e}rio, Junhyek Han, Kanishka Rao, Karl Pertsch, Karol Hausman, Keegan Go, Keerthana Gopalakrishnan, Ken Goldberg, Kendra Byrne, Kenneth Oslund, Kento Kawaharazuka, Kevin Zhang, Krishan Rana, Krishnan Srinivasan, Lawrence~Yunliang Chen, Lerrel Pinto, Liam Tan, Lionel Ott, Lisa Lee, Masayoshi Tomizuka, Maximilian Du, Michael Ahn, Mingtong Zhang, Mingyu Ding, Mohan~Kumar Srirama, Mohit Sharma, Moo~Jin Kim, Naoaki Kanazawa, Nicklas Hansen, Nicolas Heess, Nikhil~J. Joshi, Niko Suenderhauf, Norman Di~Palo, Nur Muhammad~Mahi Shafiullah, Oier Mees, Oliver Kroemer, Pannag~R. Sanketi, Paul Wohlhart, Peng Xu, Pierre Sermanet, Priya Sundaresan, Quan Vuong, Rafael Rafailov, Ran Tian, Ria Doshi, Roberto {Mart{\'i}n-Mart{\'i}n}, Russell Mendonca, Rutav Shah, Ryan Hoque, Ryan Julian, Samuel Bustamante, Sean Kirmani, Sergey Levine, Sherry Moore, Shikhar Bahl, Shivin Dass, Shubham Sonawani, Shuran Song, Sichun Xu, Siddhant Haldar, Simeon Adebola, Simon Guist, Soroush Nasiriany, Stefan Schaal, Stefan Welker, Stephen Tian, Sudeep Dasari, Suneel Belkhale, Takayuki Osa, Tatsuya Harada, Tatsuya Matsushima, Ted Xiao, Tianhe Yu, Tianli Ding, Todor Davchev, Tony~Z. Zhao, Travis Armstrong, Trevor Darrell, Vidhi Jain, Vincent Vanhoucke, Wei Zhan, Wenxuan Zhou, Wolfram Burgard, Xi~Chen, Xiaolong Wang, Xinghao Zhu, Xuanlin Li, Yao Lu, Yevgen Chebotar, Yifan Zhou, Yifeng Zhu, Ying Xu, Yixuan Wang, Yonatan Bisk, Yoonyoung Cho, Youngwoon Lee, Yuchen Cui, Yueh-Hua Wu, Yujin Tang, Yuke Zhu, Yunzhu Li, Yusuke Iwasawa, Yutaka Matsuo, Zhuo Xu, and Zichen~Jeff Cui.
\newblock Open {{X-Embodiment}}: {{Robotic Learning Datasets}} and {{RT-X Models}}, October 2023.

\bibitem[Dai et~al.(2026)Dai, Fu, Lee, Liu, Zhang, Yang, Finn, Fazeli, and Chai]{daiRoboMMEBenchmarkingUnderstanding2026}
Yinpei Dai, Hongze Fu, Jayjun Lee, Yuejiang Liu, Haoran Zhang, Jianing Yang, Chelsea Finn, Nima Fazeli, and Joyce Chai.
\newblock {{RoboMME}}: {{Benchmarking}} and {{Understanding Memory}} for {{Robotic Generalist Policies}}, May 2026.

\bibitem[Dai et~al.(2019)Dai, Yang, Yang, Carbonell, Le, and Salakhutdinov]{daiTransformerXLAttentiveLanguage2019}
Zihang Dai, Zhilin Yang, Yiming Yang, Jaime Carbonell, Quoc~V. Le, and Ruslan Salakhutdinov.
\newblock Transformer-{{XL}}: {{Attentive Language Models Beyond}} a {{Fixed-Length Context}}, June 2019.

\bibitem[Fu et~al.(2024)Fu, Huang, Datta, Chen, Panitch, Liu, Li, and Goldberg]{fuInContextImitationLearning2024}
Letian Fu, Huang Huang, Gaurav Datta, Lawrence~Yunliang Chen, William Chung-Ho Panitch, Fangchen Liu, Hui Li, and Ken Goldberg.
\newblock In-{{Context Imitation Learning}} via {{Next-Token Prediction}}, September 2024.

\bibitem[Haarnoja et~al.(2019)Haarnoja, Zhou, Hartikainen, Tucker, Ha, Tan, Kumar, Zhu, Gupta, Abbeel, and Levine]{haarnojaSoftActorCriticAlgorithms2019}
Tuomas Haarnoja, Aurick Zhou, Kristian Hartikainen, George Tucker, Sehoon Ha, Jie Tan, Vikash Kumar, Henry Zhu, Abhishek Gupta, Pieter Abbeel, and Sergey Levine.
\newblock Soft {{Actor-Critic Algorithms}} and {{Applications}}, January 2019.

\bibitem[Huang et~al.(2026)Huang, Li, Guo, Zhang, Peng, Zhou, Ruan, Wu, Wang, Su, Xie, Chen, Xu, Zhou, Chen, Zhao, He, Li, Li, Dong, Dang, Huang, Chen, Chang, Zhang, Deng, Wang, Wei, Li, Yang, Zhou, Liu, Cheng, Shao, Wang, Jin, Hao, Chen, and Li]{huang2026incontextlearningrobotsmethods}
Haojian Huang, Zexi Li, Junhao Guo, Yehang Zhang, Wenxuan Peng, Bohan Zhou, Weilin Ruan, Leyi Wu, Chenxu Wang, Jianchong Su, Binghui Xie, Wosong Chen, Yingjie Xu, Tianhao Zhou, Suzeyu Chen, Pukun Zhao, Jiaqi He, Xinyi Li, Runze Li, Peiran Dong, Shaoxiang Dang, Jing Huang, Yingbing Chen, Yifan Chang, Tianyi Zhang, Shiyuan Deng, Haozhi Wang, Yangkai Wei, Wenqian Li, Han Yang, Kaiwen Zhou, Huaping Liu, James Cheng, Rui Shao, Donglin Wang, Yaochu Jin, Jianye Hao, Ying-Cong Chen, and Yinchuan Li.
\newblock In-context learning for robots: Methods and applications, 2026.
\newblock URL \url{https://arxiv.org/abs/2609.36012}.

\bibitem[Intelligence et~al.(2025{\natexlab{a}})Intelligence, Amin, Aniceto, Balakrishna, Black, Conley, Connors, Darpinian, Dhabalia, DiCarlo, Driess, Equi, Esmail, Fang, Finn, Glossop, Godden, Goryachev, Groom, Hancock, Hausman, Hussein, Ichter, Jakubczak, Jen, Jones, Katz, Ke, Kuchi, Lamb, LeBlanc, Levine, {Li-Bell}, Lu, Mano, Mothukuri, Nair, Pertsch, Ren, Sharma, Shi, Smith, Springenberg, Stachowicz, Stoeckle, Swerdlow, Tanner, Torne, Vuong, Walling, Wang, Williams, Yoo, Yu, Zhilinsky, and Zhou]{intelligence$p^_06$VLAThat2025}
Physical Intelligence, Ali Amin, Raichelle Aniceto, Ashwin Balakrishna, Kevin Black, Ken Conley, Grace Connors, James Darpinian, Karan Dhabalia, Jared DiCarlo, Danny Driess, Michael Equi, Adnan Esmail, Yunhao Fang, Chelsea Finn, Catherine Glossop, Thomas Godden, Ivan Goryachev, Lachy Groom, Hunter Hancock, Karol Hausman, Gashon Hussein, Brian Ichter, Szymon Jakubczak, Rowan Jen, Tim Jones, Ben Katz, Liyiming Ke, Chandra Kuchi, Marinda Lamb, Devin LeBlanc, Sergey Levine, Adrian {Li-Bell}, Yao Lu, Vishnu Mano, Mohith Mothukuri, Suraj Nair, Karl Pertsch, Allen~Z. Ren, Charvi Sharma, Lucy~Xiaoyang Shi, Laura Smith, Jost~Tobias Springenberg, Kyle Stachowicz, Will Stoeckle, Alex Swerdlow, James Tanner, Marcel Torne, Quan Vuong, Anna Walling, Haohuan Wang, Blake Williams, Sukwon Yoo, Lili Yu, Ury Zhilinsky, and Zhiyuan Zhou.
\newblock \${$\pi$}\textasciicircum\textbraceleft *\textbraceright\_\textbraceleft 0.6\textbraceright\$: A {{VLA That Learns From Experience}}, November 2025{\natexlab{a}}.

\bibitem[Intelligence et~al.(2025{\natexlab{b}})Intelligence, Black, Brown, Darpinian, Dhabalia, Driess, Esmail, Equi, Finn, Fusai, Galliker, Ghosh, Groom, Hausman, Ichter, Jakubczak, Jones, Ke, LeBlanc, Levine, {Li-Bell}, Mothukuri, Nair, Pertsch, Ren, Shi, Smith, Springenberg, Stachowicz, Tanner, Vuong, Walke, Walling, Wang, Yu, and Zhilinsky]{intelligence$p_05$VisionLanguageActionModel2025}
Physical Intelligence, Kevin Black, Noah Brown, James Darpinian, Karan Dhabalia, Danny Driess, Adnan Esmail, Michael Equi, Chelsea Finn, Niccolo Fusai, Manuel~Y. Galliker, Dibya Ghosh, Lachy Groom, Karol Hausman, Brian Ichter, Szymon Jakubczak, Tim Jones, Liyiming Ke, Devin LeBlanc, Sergey Levine, Adrian {Li-Bell}, Mohith Mothukuri, Suraj Nair, Karl Pertsch, Allen~Z. Ren, Lucy~Xiaoyang Shi, Laura Smith, Jost~Tobias Springenberg, Kyle Stachowicz, James Tanner, Quan Vuong, Homer Walke, Anna Walling, Haohuan Wang, Lili Yu, and Ury Zhilinsky.
\newblock \${$\pi\_$}\textbraceleft 0.5\textbraceright\$: A {{Vision-Language-Action Model}} with {{Open-World Generalization}}, April 2025{\natexlab{b}}.

\bibitem[Jiang et~al.(2026)Jiang, Chebotar, Zheng, Hu, Ge, Wu, Dai, Reed, {Fei-Fei}, Zhu, and Fan]{jiangRoboTTTContextScaling2026}
Yunfan Jiang, Yevgen Chebotar, Ruijie Zheng, Fengyuan Hu, Yunhao Ge, Jimmy Wu, Tianyuan Dai, Scott Reed, Li~{Fei-Fei}, Yuke Zhu, and Linxi~"Jim" Fan.
\newblock {{RoboTTT}}: {{Context Scaling}} for {{Robot Policies}}, July 2026.

\bibitem[Kim et~al.(2024)Kim, Pertsch, Karamcheti, Xiao, Balakrishna, Nair, Rafailov, Foster, Lam, Sanketi, Vuong, Kollar, Burchfiel, Tedrake, Sadigh, Levine, Liang, and Finn]{kim2024OpenVLAOpenSourceVisionLanguageActionModela}
Moo~Jin Kim, Karl Pertsch, Siddharth Karamcheti, Ted Xiao, Ashwin Balakrishna, Suraj Nair, Rafael Rafailov, Ethan Foster, Grace Lam, Pannag Sanketi, Quan Vuong, Thomas Kollar, Benjamin Burchfiel, Russ Tedrake, Dorsa Sadigh, Sergey Levine, Percy Liang, and Chelsea Finn.
\newblock {{OpenVLA}}: {{An Open-Source Vision-Language-Action Model}}, September 2024.

\bibitem[Laskin et~al.(2022{\natexlab{a}})Laskin, Wang, Oh, Parisotto, Spencer, Steigerwald, Strouse, Hansen, Filos, Brooks, Gazeau, Sahni, Singh, and Mnih]{laskinIncontextReinforcementLearning2022a}
Michael Laskin, Luyu Wang, Junhyuk Oh, Emilio Parisotto, Stephen Spencer, Richie Steigerwald, D.~J. Strouse, Steven Hansen, Angelos Filos, Ethan Brooks, Maxime Gazeau, Himanshu Sahni, Satinder Singh, and Volodymyr Mnih.
\newblock In-context {{Reinforcement Learning}} with {{Algorithm Distillation}}, October 2022{\natexlab{a}}.

\bibitem[Laskin et~al.(2022{\natexlab{b}})Laskin, Wang, Oh, Parisotto, Spencer, Steigerwald, Strouse, Hansen, Filos, Brooks, Gazeau, Sahni, Singh, and Mnih]{laskinIncontextReinforcementLearning2022}
Michael Laskin, Luyu Wang, Junhyuk Oh, Emilio Parisotto, Stephen Spencer, Richie Steigerwald, D.~J. Strouse, Steven~Stenberg Hansen, Angelos Filos, Ethan Brooks, Maxime Gazeau, Himanshu Sahni, Satinder Singh, and Volodymyr Mnih.
\newblock In-context {{Reinforcement Learning}} with {{Algorithm Distillation}}.
\newblock In \emph{The {{Eleventh International Conference}} on {{Learning Representations}}}, September 2022{\natexlab{b}}.

\bibitem[Li et~al.(2025)Li, Zuo, Yu, Zhang, Yang, Zhang, Zhu, Zhang, Chen, Cui, Wang, Luo, Fan, Sun, Zeng, Pang, Zhang, Wang, Mu, Zhou, and Ding]{liSimpleVLARLScalingVLA2025}
Haozhan Li, Yuxin Zuo, Jiale Yu, Yuhao Zhang, Zhaohui Yang, Kaiyan Zhang, Xuekai Zhu, Yuchen Zhang, Tianxing Chen, Ganqu Cui, Dehui Wang, Dingxiang Luo, Yuchen Fan, Youbang Sun, Jia Zeng, Jiangmiao Pang, Shanghang Zhang, Yu~Wang, Yao Mu, Bowen Zhou, and Ning Ding.
\newblock {{SimpleVLA-RL}}: {{Scaling VLA Training}} via {{Reinforcement Learning}}, September 2025.

\bibitem[Liu et~al.(2026)Liu, Gao, Wei, Chen, Liao, Wu, Yu, and Wang]{liuWhatCanRL2026}
Jijia Liu, Feng Gao, Bingwen Wei, Xinlei Chen, Qingmin Liao, Yi~Wu, Chao Yu, and Yu~Wang.
\newblock What {{Can RL Bring}} to {{VLA Generalization}}? {{An Empirical Study}}, January 2026.

\bibitem[Luo et~al.(2025)Luo, Xu, Wu, and Levine]{luoPreciseDexterousRobotic2025}
Jianlan Luo, Charles Xu, Jeffrey Wu, and Sergey Levine.
\newblock Precise and {{Dexterous Robotic Manipulation}} via {{Human-in-the-Loop Reinforcement Learning}}, March 2025.

\bibitem[Pan et~al.(2026)Pan, Feng, Zhang, Li, Song, Qu, Wang, Li, Xiong, Chen, Liu, and Luo]{panSOPScalableOnline2026}
Mingjie Pan, Siyuan Feng, Qinglin Zhang, Xinchen Li, Jianheng Song, Chendi Qu, Yi~Wang, Chuankang Li, Ziyu Xiong, Zhi Chen, Yi~Liu, and Jianlan Luo.
\newblock {{SOP}}: {{A Scalable Online Post-Training System}} for {{Vision-Language-Action Models}}, January 2026.

\bibitem[Ross et~al.(2011)Ross, Gordon, and Bagnell]{rossReductionImitationLearning2011}
Stephane Ross, Geoffrey Gordon, and Drew Bagnell.
\newblock A {{Reduction}} of {{Imitation Learning}} and {{Structured Prediction}} to {{No-Regret Online Learning}}.
\newblock In \emph{Proceedings of the {{Fourteenth International Conference}} on {{Artificial Intelligence}} and {{Statistics}}}, pp.\  627--635. {JMLR Workshop and Conference Proceedings}, June 2011.

\bibitem[Schulman et~al.(2017)Schulman, Wolski, Dhariwal, Radford, and Klimov]{schulmanProximalPolicyOptimization2017}
John Schulman, Filip Wolski, Prafulla Dhariwal, Alec Radford, and Oleg Klimov.
\newblock Proximal {{Policy Optimization Algorithms}}.
\newblock \emph{arXiv:1707.06347 [cs]}, August 2017.

\bibitem[Shao et~al.(2024)Shao, Wang, Zhu, Xu, Song, Bi, Zhang, Zhang, Li, Wu, and Guo]{shaoDeepSeekMathPushingLimits2024}
Zhihong Shao, Peiyi Wang, Qihao Zhu, Runxin Xu, Junxiao Song, Xiao Bi, Haowei Zhang, Mingchuan Zhang, Y.~K. Li, Y.~Wu, and Daya Guo.
\newblock {{DeepSeekMath}}: {{Pushing}} the {{Limits}} of {{Mathematical Reasoning}} in {{Open Language Models}}, April 2024.

\bibitem[Vaswani et~al.(2017)Vaswani, Shazeer, Parmar, Uszkoreit, Jones, Gomez, Kaiser, and Polosukhin]{vaswaniAttentionAllYou2017}
Ashish Vaswani, Noam Shazeer, Niki Parmar, Jakob Uszkoreit, Llion Jones, Aidan~N Gomez, {\L}ukasz Kaiser, and Illia Polosukhin.
\newblock Attention is {{All}} you {{Need}}.
\newblock In \emph{Advances in {{Neural Information Processing Systems}}}, volume~30. Curran Associates, Inc., 2017.

\bibitem[Wagenmaker et~al.(2025)Wagenmaker, Nakamoto, Zhang, Park, Yagoub, Nagabandi, Gupta, and Levine]{wagenmaker2025SteeringYourDiffusionPolicyLatentSpaceReinforcement}
Andrew Wagenmaker, Mitsuhiko Nakamoto, Yunchu Zhang, Seohong Park, Waleed Yagoub, Anusha Nagabandi, Abhishek Gupta, and Sergey Levine.
\newblock Steering {{Your Diffusion Policy}} with {{Latent Space Reinforcement Learning}}, June 2025.

\bibitem[Wu et~al.(2026)Wu, Lu, Wang, Yang, Liu, Wang, Zhu, Sun, Wang, Ma, Ren, Zhang, Yu, Zhao, Zhou, Qiu, Xiong, Wang, Wang, Cheng, Li, Huang, Zhu, Shen, and Zheng]{wuPragmaticVLAFoundation2026}
Wei Wu, Fan Lu, Yunnan Wang, Shuai Yang, Shi Liu, Fangjing Wang, Qian Zhu, He~Sun, Yong Wang, Shuailei Ma, Yiyu Ren, Kejia Zhang, Hui Yu, Jingmei Zhao, Shuai Zhou, Zhenqi Qiu, Houlong Xiong, Ziyu Wang, Zechen Wang, Ran Cheng, Yong-Lu Li, Yongtao Huang, Xing Zhu, Yujun Shen, and Kecheng Zheng.
\newblock A {{Pragmatic VLA Foundation Model}}, June 2026.

\bibitem[Xu et~al.(2026)Xu, Springenberg, Equi, Amin, Esmail, Levine, and Ke]{xuRLTokenBootstrapping2026}
Charles Xu, Jost~Tobias Springenberg, Michael Equi, Ali Amin, Adnan Esmail, Sergey Levine, and Liyiming Ke.
\newblock {{RL Token}}: {{Bootstrapping Online RL}} with {{Vision-Language-Action Models}}, April 2026.

\bibitem[Yu et~al.(2025)Yu, Wang, Guo, Lin, Xu, Zang, Zhang, Wu, Zhu, Hu, Huang, Wei, Xie, Yang, Dai, Xu, Du, Wang, Fu, Shi, Liu, Chen, Liu, Liu, Li, Yang, Yang, Dai, and Wang]{yuRLinfFlexibleEfficient2025}
Chao Yu, Yuanqing Wang, Zhen Guo, Hao Lin, Si~Xu, Hongzhi Zang, Quanlu Zhang, Yongji Wu, Chunyang Zhu, Junhao Hu, Zixiao Huang, Mingjie Wei, Yuqing Xie, Ke~Yang, Bo~Dai, Zhexuan Xu, Jiakun Du, Xiangyuan Wang, Xu~Fu, Letong Shi, Zhihao Liu, Kang Chen, Weilin Liu, Gang Liu, Boxun Li, Jianlei Yang, Zhi Yang, Guohao Dai, and Yu~Wang.
\newblock {{RLinf}}: {{Flexible}} and {{Efficient Large-scale Reinforcement Learning}} via {{Macro-to-Micro Flow Transformation}}, December 2025.

\bibitem[Yuan et~al.(2026)Yuan, Dong, Liu, and Zhao]{yuanFastWAMWorldAction2026}
Tianyuan Yuan, Zibin Dong, Yicheng Liu, and Hang Zhao.
\newblock Fast-{{WAM}}: {{Do World Action Models Need Test-time Future Imagination}}?, March 2026.

\end{thebibliography}
\bibliographystyle{paper}

\appendix
\clearpage
\setcounter{table}{0}
\renewcommand{\thetable}{S\arabic{table}}
\renewcommand{\theHtable}{appendix.\arabic{table}}
\section{Appendix}
\label{sec:appendix}

\subsection{Implementation Details}
\label{app:architecture}

\subsubsection{VLRLA-bridge} 
We introduce the implementation details for our \texttt{VLRLA-bridge} variant. Its hyperparameters are reported in \Cref{tab:bridge-implementation}.

  To reduce the memory cost of cross attention to a large multi-layer KV cache extracted from the base $\pi_{0.5}$ model, we introduce a learned summarization module. Its queries attend across valid cache positions and layers, producing a compact set of
  KV summarization tokens. We then provide these summary tokens to the cross attention module of  action query tokens  from which
  the bridge extracts the information used to modify the current cache.

  For input processing, we use type-specific encoders together with learned absolute positional encodings.  Proprioceptive states and individual actions are mapped into a common token space through separate linear projections, while reward, termination and truncation use separate binary embedding tables. We use learned absolute position and index embeddings at multiple levels: summary-query token embeddings and action-query token embeddings distinguish their respective token slots, action-position embeddings encode order within each
  action chunk, and transition-position embeddings locate every token within the whole trial. Finally, a shared linear projection
  maps these position-aware tokens to the Transformer's hidden dimension.

  To translate the output of VLRLA-bridge as KV cache modifications, we pool the action-query token's outputs into a representation $u_t$ and introduce a low-rank residual adapter. For a native cache vector $x_{t,\ell,p}^{s}$,
  the adapter computes
    \begin{equation}
  \begin{aligned}
  c_{t,j,p}^{s}
  &=
  \tanh\!\left(
  W_z z_{t,j,p}^{s}
  + W_u u_t
  + e_p + e_j + e_s
  \right),\\
  \Delta z_{t,j,p}^{s}
  &=
  m_{t,p}\,(B_j^{s})^\top c_{t,j,p}^{s},\\
  z_t' &= z_t + \Delta z_t,
  \end{aligned}
  \end{equation}
    where the embeddings identify cache position, layer, and type, and $m_{t,p}$ excludes padded positions. All learned adapter parameters belong to $\theta_{\mathrm{RL}}$. This parameterization combines current cache
  content with historical context while restricting residuals to learned low-rank bases. We initialize the output bases $B_j^{s}$ to zero, yielding $z_t'=z_t$ at initialization and preserving the pretrained policy's initial behavior.
 \begin{table}[!htbp]
      \centering
      \small
      \caption{Common architecture and training settings for \texttt{VLRLA-bridge}}
      \label{tab:bridge-implementation}
      \begin{tabular}{ll}
          \hline
          \textbf{Setting} & \textbf{Value} \\
          \hline
          Transformer layers / attention heads & 6 / 8 \\
          Hidden / feedforward dimension & 512 / 1024 \\
          Number of summary / action queries & 36 / 24 \\
          Residual-adapter rank & 64 \\
          Dropout & 0.1 \\
          Residual output initialization & Zero \\
          Sequence / segment length & 48 / 6 transitions \\
          \hline
          Optimizer & AdamW \\
          Peak learning rate & $10^{-4}$ \\
          Weight decay & $0.01$ \\
          Learning-rate schedule & 10\% warmup, cosine decay \\
          Gradient clipping norm & 1.0 \\
          Global batch size & 64 \\
          \hline
      \end{tabular}

  \end{table}

\clearpage
\subsection{Learning History Construction Procedures}
\label{app:dataset}
\subsubsection{In Simulation RL Dataset}
\label{app:in-sim-rl-dataset}

We document the procedure and hyperparameters for our collection of In-simulation RL Dataset here:

\paragraph{Task and observations.}
We fine-tune $\pi_{0.5}$ with proximal policy optimization (PPO) on the
various tasks in Robotwin as shown in main text (we use \texttt{adjust\_bottle} for example here, their general configurations are similar). The embodiment is ALOHA--AgileX.

\paragraph{Policy parameterization.}
We initialize the policy from an existing $\pi_{0.5}$ checkpoint configured
for ALOHA control. The vision--language backbone remains frozen, while the
action expert and associated trainable policy parameters are optimized
without low-rank adapters. Each policy call generates a chunk of $50$
actions, each with $14$ dimensions, using five flow integration steps.
Training uses the flow-SDE sampler with a noise coefficient of $0.3$:
one integration step is selected for stochastic exploration, and the
remaining steps follow the deterministic flow update. PPO surrogate for  the flow models are calcuated based on method from the Rlinf~\citep{yuRLinfFlexibleEfficient2025}.
A value head predicts returns from pooled action-expert features; its
inputs are detached so that the value loss does not update the policy
representation.

\paragraph{Reward and optimization.}
The learning signal is a sparse success reward provided by the environment
wrapper. With unit reward scaling and relative rewards enabled, a reward
of one is assigned when an episode first becomes successful. Rewards are aggregated at the action chunk level. We estimate advantages using generalized advantage estimation
with $\gamma=0.99$ and $\lambda=0.95$, and normalize the resulting advantages
before optimization. The actor uses a clipped PPO objective with ratio
bounds $[0.8,1.2]$ and an additional dual-clipping coefficient of $3.0$.
The critic uses a clipped Huber loss with value-clipping threshold $0.2$
and Huber parameter $10$. Actor and critic losses are added with equal
weight, with no entropy bonus or reference-policy KL penalty. AdamW uses
separate learning rates of $5\times10^{-6}$ for the policy and
$10^{-4}$ for the value head.

\paragraph{Rollout and update schedule.}
Training comprises $25$ PPO iterations. Each iteration collects four
rollout rounds from $256$ parallel environments, with a wrapper horizon
of $400$ environment steps per round, corresponding to 8 action-chunk
decisions per environment. The resulting rollout tensor therefore contains
$8{,}192$ chunk slots per iteration; slots following episode termination
are excluded from the training objective by a loss mask. Each rollout
batch is reused for five optimization epochs. The global optimization
minibatch contains $2{,}048$ chunk samples, with $32$ samples per GPU
microbatch and eight gradient-accumulation steps across eight GPUs.
Training uses fully sharded data parallelism on a single node, with
offloading enabled for the actor, rollout workers, and environment workers.

\paragraph{Evaluation.}
The policy is evaluated before training and every five PPO iterations
using $128$ parallel environments and one evaluation rollout round.
Evaluation uses a separately specified seed list with fixed reset-state
identifiers and the same $400$-step wrapper horizon.  Table~\ref{tab:pi05-adjust-bottle-ppo}
summarizes the training hyperparameters.

\paragraph{Fianl Dataset.}
  Each recorded decision associates a task
  instruction, three RGB observations, and proprioceptive state with a
  $50\times14$ action chunk, accompanied by reward, termination, and
  truncation arrays.  Both successful and unsuccessful rollouts are retained.
  A complete $25$-iteration run provides a nominal capacity of $204{,}800$
  chunk slots ($25\times4\times256\times8$).

\begin{table*}[t]
  \centering
  \caption{PPO fine-tuning configuration for $\pi_{0.5}$ on
  Robotwin tasks.}
  \label{tab:pi05-adjust-bottle-ppo}
  \small
  \setlength{\tabcolsep}{8pt}
  \begin{tabular}{@{}ll@{}}
    \toprule
    Hyperparameter & Value \\
    \midrule
    Policy / embodiment & $\pi_{0.5}$ / ALOHA--AgileX \\
    Visual inputs & Head RGB and two wrist RGB views \\
    Frozen component & Vision--language backbone \\
    Action dimension / chunk length & $14$ / $50$ \\
    Flow integration steps & $5$ \\
    Training sampler / noise coefficient & Flow-SDE / $0.3$ \\
    \midrule
    PPO iterations & $25$ \\
    Training / evaluation environments & $256$ / $128$ \\
    Rollout rounds per iteration & $4$ \\
    Wrapper episode limit / rollout horizon & $400$ / $400$ steps \\
    Chunk slots per iteration (before masking) & $8{,}192$ \\
    Memory detach frequency & 12 chunks (2 episodes) \\
    Optimization epochs per iteration & $5$ \\
    Global optimization minibatch & $2{,}048$ chunk samples \\
    Microbatch per GPU & $32$ chunk samples \\
    Gradient-accumulation steps & $8$ \\
    \midrule
    Reward and log-probability aggregation & Action chunk level \\
    Advantage estimator / normalization & GAE / enabled \\
    Discount factor $\gamma$ / GAE parameter $\lambda$ & $0.99$ / $0.95$ \\
    PPO clipping parameter / dual-clipping coefficient & $0.2$ / $3.0$ \\
    Value clipping / Huber parameter & $0.2$ / $10.0$ \\
    Value-loss weight & $1.0$ \\
    \midrule
    Optimizer & AdamW \\
    Policy / value-head learning rate & $5\times10^{-6}$ / $10^{-4}$ \\
    Adam coefficients $(\beta_1,\beta_2)$ / $\epsilon$ & $(0.9,0.95)$ / $10^{-8}$ \\
    Weight decay / gradient-norm clipping & $0.01$ / $1.0$ \\
    Actor seed / base environment seed & $1234$ / $0$ \\
    \bottomrule
  \end{tabular}
\end{table*}

\clearpage
\subsubsection{Human-in-the-loop Correction Datasset}
\label{app:hitl-dataset}

 \paragraph{Dataset.}
  Our human-in-the-loop (HITL) manipulation dataset
  comprising 391 episodes organized into four ordered stages.
  The dataset covers six subtasks defined by combinations of three
  fruit categories (apple, grape, and starfruit) and two target
  object categories (bowl and plate). Each episode contains
  12 action-chunk decisions. At each decision, the stored observation
  includes an RGB image from the main camera, an RGB image from a
  wrist-mounted camera, a seven-dimensional robot state, and a
  natural-language instruction. The recorded action chunk contains
  20 seven-dimensional action vectors. The main-camera and wrist-camera
  images have resolutions of $540 \times 960$ and $480 \times 640$
  pixels, respectively. In total, the dataset contains 4,692
  action-chunk decisions and 93,840 recorded action vectors.

  Each episode additionally includes human-provided annotations
  identifying the fruit category, target object category, and binary
  task success. Both successful and unsuccessful episodes are retained.
  Table~\ref{tab:hitl-dataset} summarizes the dataset composition.
  The reported success proportions describe the recorded data and
  are not evaluation results for the trained model.

  \begin{table}[t]
      \centering
      \caption{Composition of the HITL dataset. Success denotes the
      binary human annotation associated with each recorded episode.}
      \label{tab:hitl-dataset}
      \begin{tabular}{lrrr}
          \toprule
          Stage & Episodes & Successful & Success (\%) \\
          \midrule
          0 &  99 &  35 & 35.35 \\
          1 &  95 &  53 & 55.79 \\
          2 &  99 &  80 & 80.81 \\
          3 &  98 &  91 & 92.86 \\
          \midrule
          Total & 391 & 259 & 66.24 \\
          \bottomrule
      \end{tabular}
  \end{table}

  \paragraph{Construction of training sequences.}
  Let $\mathcal{D}_s$ denote the episode pool for stage
  $s \in \{0,1,2,3\}$, and let $\mathcal{D}_{s,c}$ denote its
  subset associated with subtask $c$. A training trail is formed
  by independently sampling one episode from each stage and
  concatenating the sampled episodes in stage order. Each trial
  therefore contains $4 \times 12 = 48$ action-chunk transitions.
  Sampling is performed with replacement; constructed sequences
  need not correspond to matched episodes or a single continuous
  physical trajectory. We retain the complete episode pool at
  every stage, without dataset subsampling or success-based filtering.

  \paragraph{Sampling protocols.}
  We  perform \emph{same-subtask sampling}, we first
  select one of the six subtasks uniformly and then independently
  sample an episode of that subtask from each stage:
  \begin{equation}
      c \sim \operatorname{Uniform}(\mathcal{C}),
      \qquad
      \tau_s \mid c
      \sim \operatorname{Uniform}(\mathcal{D}_{s,c}),
      \qquad |\mathcal{C}|=6.
  \end{equation}
  This protocol preserves subtask identity throughout a sequence.

  \paragraph{Preprocessing and training objective.}
  Images, instructions, states, and action targets are processed
  using the input transformations associated with the pretrained
  $\pi_{0.5}$ checkpoint. State and action preprocessing uses the
  checkpoint's quantile-normalization statistics. The \texttt{VLRLA-bridge}
  consumes seven-dimensional normalized states and action
  histories, while action targets supplied to the frozen action
  expert are padded to its 32-dimensional action representation.

  \begin{table}[t]
      \centering
      \small
      \caption{Hyperparameters for fruit-placement tasks.}
      \label{tab:training-hyperparameters}
      \begin{tabular}{@{}ll@{}}
          \toprule
          Hyperparameter & Value \\
          \midrule
          Backbone & Frozen $\pi_{0.5}$ \\
          Trainable module & Temporal residual-KV bridge \\
          Episodes per sequence & 4 (one per stage) \\
          Transitions per episode & 12 \\
          Transitions per sequence & 48 \\
          Action chunk length & 20 \\
          State / recorded action dimension & 7 / 7 \\
          \midrule
          Transformer-XL layers & 6 \\
          Hidden dimension & 512 \\
          Attention heads & 8 \\
          Feed-forward dimension & 1,024 \\
          Dropout & 0.1 \\
          Action queries & 24 \\
          KV adapter rank & 64 \\
          Segment length & 6 transitions \\
          Memory detach frequency & 12 chunks (2 episodes) \\
          Segments per optimizer update & 8 per sequence \\
          \midrule
          Training objective & Flow-matching velocity MSE \\
          Optimizer & AdamW \\
          Peak learning rate & $10^{-4}$ \\
          Adam coefficients $(\beta_1,\beta_2)$ & $(0.9,0.999)$ \\
          Weight decay & 0.01 \\
          Gradient clipping norm & 1.0 \\
          Learning-rate schedule & Linear warmup, cosine decay \\
          Warmup & $10\%$ \\
          Global batch size & 64 \\
          Batch size per GPU & 8 \\
          Numerical precision & BF16 mixed precision \\
          Optimizer updates per run & 1,680 \\
          \bottomrule
      \end{tabular}
  \end{table}

\clearpage
\subsection{Cross-episode Memory Dataset Assembly}
\label{app:mem_data}

\subsubsection{General Procedure}
We construct training sequences for the cross-episode memory task in \Cref{sect:rw_2} by applying \emph{hindsight relabeling} to an existing dataset of successful single-episode demonstrations. No additional data collection under ambiguous instructions is required.

Each source demonstration follows a color-specific instruction, such as \textit{``put the power adapter into the green plate''} or \textit{``put the power adapter into the yellow plate''}, which identifies its achieved destination color. We retain the recorded observations, robot states, and actions, but replace every instruction with the underspecified command \textit{``put the power adapter into the plate''}.

To assemble a trial of $K$ episodes, we sample an initial demonstration with destination color $c_o$ and assign binary feedback $b \in \{0,1\}$. This feedback determines the \textit{latent} target color $c_n$: if $b=1$, the target is $c_o$; if $b=0$, the target is the alternative color. Neither color is explicitly supplied to the policy; the instruction remains underspecified. For each episode $i=0,\ldots,K-1$, let $c_o^{(i)}$ denote its original destination color, with $c_o^{(0)}=c_o$. We relabel its terminal reward as
\begin{equation}
r_{c_o^{(i)}}(c_n) =
\begin{cases}
1, & c_o^{(i)}=c_n \quad \text{(match)},\\
0, & c_o^{(i)}\neq c_n \quad \text{(mismatch)}.
\end{cases}
\end{equation}
We sample the remaining $K-1$ demonstrations from successful episodes reaching $c_n$. The resulting reward sequence is $(b,1,\ldots,1)$: the initial attempt may match or mismatch the target color, while all subsequent episodes demonstrate matching behavior.

We also construct counterfactual pairs with the same initial demonstration but opposite initial feedback. The member with $b=1$ continues with demonstrations reaching the original color, while the member with $b=0$ continues with demonstrations reaching the alternative color. Both members use the same ambiguous instruction and the same augmentation randomness for the initial history. Thus, the same initial attempt leads to different subsequent action targets depending on the feedback, encouraging the policy to use feedback stored in memory to resolve the instruction's ambiguity.

The dataset contains 185 successful source episodes. We hold out 16 episodes (8 per color) for validation and dynamically assemble training trials from the remaining 169. We apply a curriculum over trial length $K$: $K=2$ for the first third of training updates, followed by alternating $K=2$ and $K=6$ for the remaining two-thirds. This curriculum trains immediate adaptation to feedback and retention of the inferred color across subsequent episodes.

\subsubsection{Detailed Settings}

  \paragraph{Real-world manipulation dataset.}
  We use a real-world Franka robot arm and a dataset comprising
  185 successful demonstrations of placing a power adapter into
  either a green or a yellow plate. The dataset contains 93
  green-target and 92 yellow-target demonstrations. Each episode
  contains 180 action steps recorded at 15\,Hz, corresponding to
  12\,s of recorded interaction. The dataset therefore comprises
  33{,}300 action steps, or approximately 37\,min of interaction.
  Observations include three RGB camera views (front, side, and
  wrist) and a 15-dimensional robot state. Actions are represented
  by seven-dimensional end-effector delta commands, comprising
  translation, rotation, and gripper control. 

  \paragraph{Temporal representation and preprocessing.}
  Each episode is partitioned into six non-overlapping chunks of
  30 actions. For each chunk, we retain the three camera images
  and robot state at its initial timestep, yielding observation
  indices $\{0,30,60,90,120,150\}$. Thus, each episode provides
  six observation-conditioned action targets, each spanning
  2\,s. RGB images are resized from $480 \times 640$ to
  $224 \times 224$ using aspect-ratio-preserving padding.
  Preprocessing normalizes the end-effector quaternion to unit
  length and resolves its sign ambiguity by making its
  largest-magnitude component nonnegative. Gripper state and
  action conventions are aligned with the pretrained policy.
  State and action normalization uses the quantile statistics
  bundled with the frozen base-model checkpoint.

  \begin{table*}[t]
  \centering
  \small
  \caption{Dataset specification and adapter-training
  hyperparameters.  }
  \label{tab:dataset_training_settings}
  \begin{tabularx}{0.85\textwidth}{@{}l>{\raggedright\arraybackslash}X@{}}
  \toprule
  Parameter & Value \\
  \midrule
  \multicolumn{2}{@{}l}{\textit{Dataset and representation}} \\
  Successful source episodes & 185: 93 green, 92 yellow \\
  Adapter-training episodes & 169: 85 green, 84 yellow \\
  Held-out source episodes & 16: 8 per color \\
  Recording frequency & 15\,Hz \\
  Episode length & 180 action steps (12\,s) \\
  Observation views & Front, side, wrist RGB \\
  Source / processed image size
      & $480 \times 640$ / $224 \times 224$ \\
  Robot-state / action dimension & 15 / 7 \\
  Action horizon & 30 steps per chunk \\
  Chunks per episode & 6 \\
  Training-trail length & 2 or 6 episodes \\
  Normalization & Frozen checkpoint quantile statistics \\
  \midrule
  \multicolumn{2}{@{}l}{\textit{Architecture}} \\
  Base policy & Frozen $\pi_{0.5}$ \\
  Transformer layers & 6 \\
  Model / feed-forward width & 512 / 1{,}024 \\
  Summary / adaptation queries & 72 / 48 \\
  Residual adapter rank & 64 \\
  Memory detach frequency & 12 chunks (2 episodes) \\
  \midrule
  \multicolumn{2}{@{}l}{\textit{Optimization and curriculum}} \\
  Optimizer & AdamW, $(\beta_1,\beta_2)=(0.9,0.999)$ \\
  Weight decay & 0.01 \\
  Peak learning rate & $10^{-4}$ \\
  Optimizer updates & 2{,}790 \\
  Global / per-GPU batch size & 64 / 8 trails \\
  Numerical precision & BF16 \\
  Learning-rate warmup & 4\% of training \\
  Learning-rate decay & Cosine decay after the 60\% boundary to zero \\
  Short-trail stage & Updates 1--1{,}256: 2 episodes \\
  Mixed-trail stage & Updates 1{,}257--2{,}790: alternating 2 and 6 episodes \\
  \midrule
  \multicolumn{2}{@{}l}{\textit{Augmentation and evaluation}} \\
  Front-view crop & 95\% of each spatial dimension \\
  Front-view rotation & Uniform in $[-5^\circ,5^\circ]$ \\
  Brightness multiplier
      & $[0.85,1.15]$ \\
  Contrast multiplier
      & $[0.8,1.2]$  \\
  Saturation multiplier
      & $[0.8,1.2]$ \\
  \bottomrule
  \end{tabularx}
  \end{table*}

\clearpage
\subsection{FastWAM}
\label{app:fastwam}

A possible explanation for FastWAM's lower performance on \texttt{place\_empty\_cup} in \Cref{sec:sim-base-models} is the representation extracted to downstream policy module. Native FastWAM predicts actions using  video key/value representations from multiple backbone layers. For simplicity, our integration implementaion instead conditions on the final video hidden states. Consequently, \texttt{VLRLA-simp} does
  not retain native FastWAM's access to intermediate representations.

  The encoder also differs substantially in prefix length compared to $\pi_{0.5}$. FastWAM produces 120 visual-language tokens of dimension 3,072, produced by VAE encoding and DiT patching of the combined three-camera image. In comparison,
  $\pi_{0.5}$ receives a prefix of 968 tokens of dimension 2,048. Hence FastWAM's coarser spatial tokenization may make fine geometric
  relationships less accessible to the downstream policy module.

  These differences may disproportionately affect \texttt{place\_empty\_cup}, which requires accurate cup--coaster alignment during grasping, transport, and placement, often under gripper occlusion. Such behavior may
  benefit from spatial detail and intermediate features that are less accessible from the FastWAM prefix we used. In contrast, \texttt{adjust\_bottle} may be more tolerant of this representation choice. We therefore consider the representation interface and spatial compression plausible contributors to the observed gap.

\end{document}